\pdfoutput=1

\documentclass[11pt]{article}

\usepackage[]{emnlp2021}

\usepackage{times}
\usepackage{latexsym}

\usepackage[T1]{fontenc}

\usepackage[utf8]{inputenc}
\usepackage{array}

\usepackage{microtype}
\usepackage{linguex}

\definecolor{cedarchest}{rgb}{0.79,0.35,0.29}
\definecolor{airsuperiorityblue}{rgb}{0.45,0.63,0.76}
\definecolor{aero}{rgb}{0.49,0.73,0.91}
\definecolor{africanviolet}{rgb}{0.7,0.52,0.75}
\definecolor{amaranthpink}{rgb}{0.95,0.61,0.73}
\definecolor{absolutezero}{rgb}{0.0,0.28,0.73}

\definecolor{bluegold1}{rgb}{1.0, 0.7, 0.0}
\definecolor{bluegold2}{rgb}{0.8, 0.62, 0.25}
\definecolor{bluegold3}{rgb}{0.59, 0.54, 0.5}
\definecolor{bluegold4}{rgb}{0.42, 0.47, 0.75}
\definecolor{bluegold5}{rgb}{0.25, 0.4, 0.96}

\definecolor{pinkgold1}{rgb}{0.9, 0.17, 0.31}
\definecolor{pinkgold2}{rgb}{0.91, 0.25, 0.25}
\definecolor{pinkgold3}{rgb}{0.925, 0.33, 0.20}
\definecolor{pinkgold4}{rgb}{0.94, 0.43, 0.16}
\definecolor{pinkgold5}{rgb}{0.95, 0.51, 0.12}
\definecolor{pinkgold6}{rgb}{0.96, 0.60, 0.08}
\definecolor{pinkgold7}{rgb}{0.98, 0.65, 0.04}
\definecolor{pinkgold8}{rgb}{1.0, 0.7, 0.0}

\definecolor{ruddy}{rgb}{1.0, 0.0, 0.16}
\definecolor{rose}{rgb}{1.0, 0.0, 0.5}
\definecolor{razzledazzlerose}{rgb}{1.0, 0.2, 0.8}
\definecolor{bittersweet}{rgb}{1.0, 0.44, 0.37}
\definecolor{coral}{rgb}{1.0, 0.5, 0.31}
\definecolor{deepsaffron}{rgb}{1.0, 0.6, 0.2}
\definecolor{uclagold}{rgb}{1.0, 0.7, 0.0}
\definecolor{mustard}{rgb}{1.0, 0.86, 0.35}
\definecolor{amethyst}{rgb}{0.6, 0.4, 0.8}

\definecolor{usccardinal}{rgb}{0.6, 0.0, 0.0}
\definecolor{zaffre}{rgb}{0.0, 0.08, 0.66}

\definecolor{lime(web)(x11green)}{rgb}{0.0, 1.0, 0.5}
\definecolor{lime(colorwheel)}{rgb}{0.1, 0.9, 0.5}
\definecolor{electricyellow}{rgb}{0.2, 0.8, 0.5}
\definecolor{canaryyellow}{rgb}{0.3, 0.7, 0.5}
\definecolor{peridot}{rgb}{0.4, 0.6, 0.5}
\definecolor{schoolbusyellow}{rgb}{0.5, 0.5, 0.5}
\definecolor{yellow(munsell)}{rgb}{0.6, 0.4, 0.5}
\definecolor{tangerineyellow}{rgb}{0.7, 0.3, 0.5}
\definecolor{uclagoldfake}{rgb}{0.8, 0.2, 0.5}
\definecolor{chromeyellow}{rgb}{0.9, 0.1, 0.5}

\usepackage{pgfplots}
\usetikzlibrary{pgfplots.groupplots}
\pgfplotsset{compat=1.3}
\usepackage{tikz}
\usetikzlibrary{patterns}
\usepackage{amsfonts}
\usepackage{amssymb}
\usepackage{amsthm}
\usepackage{enumerate}
\usepackage{enumitem}
\usepackage{booktabs}
\theoremstyle{plain}

\usepackage{adjustbox}
\usepackage{xfrac}
\usepackage{breqn}
\usepackage{scalerel}
\usepackage{subfig}
\usepackage{xspace}
\usepackage{url}
\usepackage{graphicx}
\usepackage{multirow}
\usepackage{pgfplots}
\usepackage{algorithm}
\usepackage{fixltx2e}

\usepackage[noend]{algpseudocode}
\pgfplotsset{
tick label style={font=\small},
label style={font=\small},
title style={font=\normalsize},
legend style={font=\footnotesize}
}

\usepackage{transparent}
\usepackage{xcolor}
\usepackage{efbox}
\usepackage{multirow}

\usepackage{cleveref}
\crefname{section}{\S}{\S\S}
\Crefname{section}{\S}{\S\S}
\crefname{table}{Table}{Tables}
\crefname{figure}{Figure}{Figures}
\crefname{algorithm}{Algorithm}{}
\crefname{equation}{eq.}{}
\crefname{appendix}{Appendix}{}
\crefformat{section}{\S#2#1#3}

\usepackage{todonotes}
\makeatletter
\newcommand*\iftodonotes{\if@todonotes@disabled\expandafter\@secondoftwo\else\expandafter\@firstoftwo\fi} 
\makeatother

\newcommand{\calS}{\mathcal{S}}
\newcommand{\calP}{\mathcal{P}}

\newtheorem*{reh}{\textit{Reduce Error Hypothesis}}
\newtheorem*{ceh}{\textit{Cause Error Hypothesis}}

\title{Frequency Effects on Syntactic Rule Learning in Transformers}

\author{Jason Wei$^1$ \hspace{4mm} Dan Garrette$^1$ \hspace{4mm} Tal Linzen$^{2}$\thanks{\hspace{1.3mm} Work done while visiting Google.} \hspace{4mm} Ellie Pavlick$^{1,3}$ \\
  $^1$Google Research \hspace{3mm} $^2$New York University \hspace{3mm} $^3$Brown University \\
  \texttt{\{jasonwei,dhgarrette,linzen,epavlick\}@google.com} \\}

\begin{document}

\setlength{\abovedisplayskip}{3pt}
\setlength{\belowdisplayskip}{3pt}

\maketitle
\begin{abstract}
Pre-trained language models perform well on a variety of linguistic tasks that require symbolic reasoning, raising the question of whether such models implicitly represent abstract symbols and rules.
We investigate this question using the case study of BERT's performance on English subject--verb agreement. Unlike prior work, we train multiple instances of BERT from scratch, allowing us to perform a series of controlled interventions at pre-training time.
We show that BERT often generalizes well to subject--verb pairs that never occurred in training, suggesting a degree of rule-governed behavior. 
We also find, however, that performance is heavily influenced by word frequency, with experiments showing that both the absolute frequency of a verb form, as well as the frequency relative to the alternate inflection, are causally implicated in the predictions BERT makes at inference time. 
Closer analysis of these frequency effects reveals that BERT's behavior is consistent with a system that correctly applies the SVA rule in general but struggles to overcome strong training priors and to estimate agreement features (singular vs.\ plural) on infrequent lexical items.  
\end{abstract}

\section{Introduction}
Many natural language phenomena are best described as the product of applying rules to abstract symbols, without access to the content of these symbols \cite{smolensky1988proper,fodor1988connectionism}. 
Most speakers of English will agree, for example, that if ``\emph{gorp}'' is a singular noun, then, regardless of the meaning of ``\emph{gorp}'', the utterance \mbox{``\emph{the gorp \textbf{\underline{adds}} nothing}''} is grammatical, but \mbox{``\emph{the gorp \textbf{\underline{add}} nothing}''} is not. 

The success of contemporary neural language models such as BERT \cite{devlin2018bert} on language understanding tasks, as well as in more targeted linguistic evaluations \cite{marvin2018targeted,goldberg2019assessing}, raises the question of whether these systems acquire such symbolic rules. 
While previous studies have attempted to address such questions, particularly in relation to BERT \cite{rogers-etal-2020-primer}, prior work has generally not analyzed the relationship between the model's pre-training data and its behavior.
As a result, it has been difficult to tease apart the many factors that may influence a model's test time performance. 

In this paper, we investigate whether pre-trained transformer-based language models learn and apply symbolic rules, focusing on BERT's ability to follow the English subject--verb number agreement rule (\cref{sec:background}) as a case study. 
On our evaluation stimuli (\cref{sec:setup}), we find that BERT achieves high performance, even on subject--verb pairs that never occurred together in the training set (\cref{subsec:overall_performance}--\cref{subsec:label_unseen_sv_pairs}).
In exploratory data analysis, however, we find that this performance is also influenced by effects from both absolute and relative frequency of verb forms in the training data (\cref{subsec:raw_frequency}--\cref{subsec:ratio_frequency}).
To confirm these phenomena causally, we perform a series of training interventions where we pre-train BERT models on training data for which we have carefully manipulated the frequencies of verb forms (\cref{sec:manipulation}).
We further use probing classifiers to attribute observed mistakes either to errors in rule-following or to errors in lexical categorization (\cref{sec:probe}).

These experiments reveal several insights about BERT in the context of rule-governed tasks. 
First, the high performance of BERT on subject--verb combinations that never occurred in the training set is consistent with a model that learns abstract representations of lexical items and patterns, i.e., abstract features and rules.
Second, BERT's performance is influenced by absolute frequency effects, but probing classifiers show that this influence can be explained by the model's inability to learn the features of a verb form (singular vs.\ plural) for infrequent lexical items, rather than a failure to apply the rule when the verb form has been classified.
Finally, although BERT generally applies rules with high accuracy, it fails to overcome strong priors during training---when one verb form is much more frequent than another, BERT tends to produce the more common form, even when it is not consistent with the rule.

\section{Experimental Logic}
\label{sec:logic}

\subsection{Hypotheses}
We aim to investigate BERT's ability to reason over abstract symbols. 
As a case study, we focus on subject--verb agreement (SVA) in English, for which the grammaticality rule of interest is:
\begin{align*}
\textsc{number}(\textit{subject})=\textsc{number}(\textit{verb})
\end{align*}

\noindent We consider three alternative hypotheses about the process underlying BERT's behavior on SVA.

\paragraph{H1: Idealized Symbolic Learner.} 
In theory, symbolic reasoners operate over abstract categories, such as the \textit{agreement feature} \textsc{number}, and rules, such as \textit{``if} \mbox{\textsc{number}(\textit{subject}) $=$ \textsc{singular}}, \textit{then} \mbox{\textsc{number}(\textit{verb}) $=$ \textsc{singular}.''} 
Early work \citep{fodor1988connectionism} which discusses the behavior of such symbolic systems often presents an idealized version, for the sake of theoretical argument. Thus, under H1, this system would not make errors such as misclassifying inputs or erroneously parsing the sentence, and is not affected by word-specific properties (e.g., frequency).

\paragraph*{H2: Item-Specific Learner.} 
The antithesis of the idealized symbolic learner is a model that reasons entirely using word co-occurrences. This system does not represent any abstractions over the immediate inputs it receives, and thus cannot reason over features such as singular/plural. Conceptually, it is analogous to early phrase-based MT systems \citep{brown-etal-1990-statistical} that build a literal string look-up table in order to predict the most likely output given an input. By definition, it performs poorly on noun–verb pairs that never co-occurred in training, as the lookup table will not have the relevant entry.\footnote{We do not specify whether such a model has access to abstract features other than agreement because such features (e.g., the notion of subject) do not affect the specific hypotheses we consider. For example, a model that does not represent agreement feature and only learns word co-occurrences will perform poorly on unseen items, regardless of whether it has access to correct parses.} 

\paragraph*{H3: Symbolic Learner with Noisy Observations.} 
Both H1 and H2 represent extreme, largely theoretical models of system behavior. In practice, we expect systems like BERT to display some hybrid of the two. However, to our knowledge, there has been no work to date which proposes a specific hypothesis of what type of hybridization best explains BERT's behavior. In this work, we consider one such hybrid: a system that is symbolic at its core but has noisy observations.\footnote{Here, ``observations'' involves both the parser as well as the lexicon. I.e., H3 allows for errors to arise due to incorrect lexical entries and/or incorrect parses. However, since our experiments (\cref{sec:probe}) don't differentiate lexicon errors from parse errors, we do not differentiate them within this hypothesis. Future work that differentiates these two errors sources could be worthwhile.} That is, under H3, the system represents symbols (e.g. singular/plural word categories) and rules (e.g., SVA) correctly, but can make errors in mapping from inputs to symbols. Conceptually, it is analogous to a BayesNet \citep{pearl1988probabilistic} that correctly represents nodes and causal connections internally but may nonetheless incorrectly process an input, activating the wrong nodes and thus producing the wrong output. 
Thus, unlike in H1, systems consistent with H3 make errors when they cannot identify whether a subject or verb is singular or plural, potentially due to frequency effects \citep[present at all levels of processing; ][]{marantz2013words}. 

\subsection{Predictions and Summary of Findings}

We use three diagnostics to differentiate the above hypotheses: (1) generalization to unseen noun-verb pairs, (2) the presence of frequency effects when making predictions for seen noun-verb pairs, (3) and correlation between specific types of errors.

{\bf Generalization to unseen noun-verb pairs} allows us to differentiate H2 from H1 and H3. For instance, since whether the sentence \textit{``the section adds nothing''} obeys the SVA rule depends only on \mbox{\textsc{number}(\textit{``section''})} and \mbox{\textsc{number}(\textit{``adds''})}, a symbolic reasoner's ability to assess grammaticality should not depend on how frequently the words \textit{``section''} and \textit{``adds''} have been seen together in the data. Instead, we would expect such a system to learn the correct agreement features of the two words independently and apply a general SVA rule to them. In contrast, an item-specific learner, which does not represent abstract agreement features, would rely on probabilities defined over specific lexical items, and thus may fail to reason correctly about rare or unseen situations, for which such probabilities are poorly calibrated.

The {\bf presence of frequency effects} in BERT's performance allows us to differentiate H1 from H2 and H3. That is, under both H2 and H3, the model may perform worse on less frequent words (albeit for different reasons). In contrast, a system consistent with H1 should not exhibit any differences in performance due to differences in inputs below the abstraction of singular/plural.

Our experiments show that BERT generalizes well (though not perfectly) to unseen noun-verb pairs (\cref{subsec:overall_performance}--\cref{subsec:label_unseen_sv_pairs}) and exhibits clear frequency effects (\cref{sec:manipulation}). Together, these results are most consistent with a hybrid system like H3. To confirm this, we use probing classifiers to investigate H3's specific prediction about {\bf correlations between types of errors}, i.e., that errors on SVA should be explained by errors in classifying singular vs.\ plural (\cref{sec:probe}). We find that the expected error patterns explain some frequency effects (those due to absolute frequency) but not others (those due to relative frequency). Thus, we ultimately conclude that, of the hypotheses considered, H3 is the best model of BERT's behavior, though BERT exhibits additional sensitivity to frequency imbalances between competing word forms that H3 leaves underspecified.

\section{Related Work}\label{sec:background}

\paragraph{Targeted Syntactic Evaluation.}

We use the targeted syntactic evaluation framework of \citet{linzen-etal-2016-assessing} and \citet{marvin2018targeted} to measure the model's ability to learn and apply the SVA rule. Following the setup from \citet{goldberg2019assessing}, each test instance consists of a sentence in which a verb has been masked out, and BERT's masked language modeling (MLM) parameters are used to score whether the singular or plural form of the verb is a better fit for the masked position. For example, given the sentence \textit{``The section \texttt{[MASK]} nothing to the info.''} and set of verb inflections \{\textit{``add''}, \textit{``adds''}\}, the model would be considered correct if the MLM prediction assigns a higher score to the singular form \textit{``adds''} than the plural form \textit{``add''} since the subject of the masked verb position is \textit{``section,''} which is singular. 

Due to the particulars of BERT's MLM task setup, the model is only able to score words that are represented by a single wordpiece. While \citet{goldberg2019assessing} dealt with this limitation by restricting evaluation to just those verbs that appear in the original BERT model's vocabulary as a single wordpiece, we are able to avoid such compromises because pre-training the models ourselves means that we can add any entries we want to the vocabulary.

\begin{table*}[ht]
    \centering
    \small
    \begin{tabular}{cl}
    Natural & {\bf Addition} of such minor characters {\bf seem/seems} more promotional than encyclopedic. \\
    & Other popular trade {\bf items} of the area {\bf include/includes} sandalwood, rubber, and teak. \\
    & The {\bf party} that originally buys the securities effectively {\bf act/acts} as a lender. \\
    \midrule
    Nonce & The {\bf astronomer} of the first session of the 
    court during that year {\bf perform/performs} a...\\
    & The {\bf isometry} in the gulf {\bf market/markets} santa catalina island. \\
    & The {\bf sheepdog} of basic needs providers ... %
    {\bf review/reviews} a damaging effect.\\
    \end{tabular}
    \vspace{-1.5mm}
    \caption{Examples of natural and nonce stimuli. Target verbs and their subjects are bolded. The model takes as input the sentence with the verb masked, and is evaluated 
    based on which verb inflection it scores more highly.
    }
    \label{tab:example-stimuli}
    \vspace{-1.5mm}
\end{table*}

\paragraph{Syntactic Reasoning in LMs.}
There has been substantial prior work on the ability of language models to perform abstract syntactic processing tasks \cite{hu-etal-2020-systematic} (see \citet{linzen2020syntactic} for a review). 
On SVA specifically, \citet{goldberg2019assessing} found that BERT achieves high accuracy on both natural sentences (97\%) and nonce sentences (83\%), and that error rate was independent of the number of ``distractor'' words between the subject and verb; \citet{yu-etal-2020-word} showed that language models do not exhibit better grammatical knowledge of more frequent nouns.
Other work has found that BERT's performance is sensitive to factors that may suggest item-specific learning; \citet{chaves-2021-look} found that BERT's performance on number agreement is sensitive to the verb, across seven different verbs, and \citet{newman-2021-refining} found that language models performed better on verbs that they predicted were likely in context. 
The focus on frequency effects also relates to a more general line of work on understanding the effect of training size and distribution on neural language models' generalization \citep{warstadt-etal-2020-learning,lovering2021predicting}. 
To our knowledge, our present study is the first to investigate these questions via controlled interventions on the model's pre-training data, making it possible to draw stronger conclusions.

Our formulation of the SVA task also relates to work which investigates neural networks' abilities to learn lexical abstractions \citep{chronis-erk-2020-bishop,kim2021testing} and to reason systematically \citep{lake2018generalization,yanaka-etal-2019-neural,kim-linzen-2020-cogs,goodwin-etal-2020-probing}. 
These studies on systematicity, however, run controlled experiments by training small models on toy data. 
Our work studies the widely-used BERT model, trained on real data and at scale.

\section{Experimental Setup}\label{sec:setup}

\subsection{Model}
Differentiating between the hypotheses presented in \cref{sec:logic} requires analyzing model performance on individual items as a function of frequencies in the training data. 
The original BERT model was trained on both English Wikipedia and BooksCorpus \cite{zhu2015aligning}. 
However, BooksCorpus is not publicly available \cite{documentationdebt21}, so when we pre-train our BERT models, we use only the Wikipedia data (2.3 billion tokens).
Despite this difference in training data, our models achieve performances comparable to the public BERT-Base release on GLUE \cite{wang-etal-2018-glue} (see \cref{appendix:bert}).

\subsection{Evaluation Stimuli}
We evaluate the model's SVA ability on two classes of stimuli: (1) \textit{natural} sentences, which are generally both syntactically and semantically coherent, and (2) \textit{nonce} sentences (following \citealt{gulordava-etal-2018-colorless}) which are grammatically valid but not necessarily semantically coherent \cite[``\emph{colorless green ideas sleep furiously}'',][]{chomsky1956three}.
Examples of each are shown in \cref{tab:example-stimuli}.

Evaluating on natural sentences provides a measurement of how well the model can be expected to perform in realistic settings, but these sentences are not ideal for a targeted SVA evaluation since they often contain additional cues relevant to verb inflection, such as other plural verbs or plural determiners, as in ``\textit{\underline{two}} \texttt{\small [SUBJECT]} \textit{and \underline{their} dogs} \texttt{\small [VERB]},'' making it difficult to discern whether a model has chosen a particular verb inflection based on the subject.
In contrast, performance on synthetic nonce sentences allows us to ensure that the only source of information about the verb's correct inflection is the subject itself.

\paragraph{Natural Stimuli.}
Following \citet{goldberg2019assessing}, for natural stimuli we use the dataset from \citet{linzen-etal-2016-assessing}, which comprises 23,298 sentences from Wikipedia.
The target verb is plural in 16,232 of these sentences and singular in 7,064 of these sentences. 
These evaluation sentences span 176 verb lemmas and 329 verb forms. 

\paragraph{Nonce Stimuli.}\label{subsubsec:nonce_stimuli}
For our nonce stimuli, we compiled a list of 200 nouns, 336 verbs, and 56 \textit{sentential contexts}---sentence templates where we remove the original subject and verb---such that any given (noun, verb, sentential context) triplet yields a grammatically correct nonce sentence. E.g., given the sentence ``\textit{the investigation of chaperones has a long history}", we can create a sentential context: ``\textit{the} \texttt{\small [SUBJECT]} \textit{of chaperones} \texttt{\small [VERB]} \textit{a long history.}''
We can then randomly chose a noun and verb from our noun and verb lists (e.g., \textit{cities} and \textit{play}) to construct a nonce sentence: ``\textit{The \underline{cities} of chaperones \underline{play} a long history.}"
Considering all possible combinations of nouns, verbs, inflections, and contexts yields a dataset of 7,526,400 sentences which is is both large (c.f., 383 sentences in \citet{gulordava-etal-2018-colorless}) and balanced in terms of number form (50\% singular and 50\% plural).

To ensure that constructed sentences are grammatically correct, we apply several manual filters (e.g., removing verbs that have ambiguous inflections), which are described in detail in \cref{appendix:non_dataset_details} (with a list of all nouns, verbs, and sentential contexts in \cref{appendix:raw_sva_dataset}). 
To verify the quality of the resulting stimuli, one of the authors manually examined 154 randomly generated nonce sentences in the same way that they would be presented to the model. 
The verb inflection was correctly predicted in all but one of the instances (with the single error attributed to carelessness), and the annotator confirmed that all generated sentences were grammatically correct. 
Our stimuli and code are available at {\small \url{https://github.com/google-research/language/tree/master/language/bertology/frequency_effects}}.

\section{Exploratory Analyses: What Factors Correlate with Error Rates?}\label{sec:freq_effects}

We first perform an exploratory analysis of how the model's abilities on the SVA task vary as a function of pre-training frequency. 
As discussed in \cref{sec:logic}, we consider generalization to unseen subject-verb pairs to be evidence of symbolic reasoning (H1 or H3), and strong frequency effects to suggest item-specific learning (H2 or H3).
Note that in these experiments, it is not the individual lexical items---the subject and verb---that are unseen, only the combination of them in a single sentence. Therefore, this analysis evaluates the model's ability to perform abstract reasoning about individual items for which it has learned representations.

\subsection{Overall Performance}\label{subsec:overall_performance}

Overall, the model's error rate is 3.2\% on natural stimuli and 16.8\% on nonce stimuli.
This is similar to \citet{goldberg2019assessing}'s reported 3\% error on natural stimuli from \citet{linzen-etal-2016-assessing} and a 17\% error on nonce stimuli from \citet{gulordava-etal-2018-colorless}.\footnote{The \citet{gulordava-etal-2018-colorless} stimuli slightly differ in that all content words (not just the subject and verb) were replaced.} %

\begingroup
\setlength{\tabcolsep}{4pt}
\newcommand{\sentry}[2]{#1\% \tiny{(#2)}}
\begin{table}[t]
    \centering
    \small
    \begin{tabular}{l *{4}{>{\raggedleft\arraybackslash}p{9.5mm}}} %
    \toprule
     & \multicolumn{2}{c}{Natural} & \multicolumn{2}{c}{Nonce} \\
     \cmidrule(lr){2-3}\cmidrule(lr){4-5}
     & Seen & Unseen & Seen & Unseen \\
    \midrule
    argmax\textsubscript{V} P(V) & 39.1 & 39.0 & 50.0 & 50.0 \\
    argmax\textsubscript{V} P(SV) & 22.9 & 50.0 & 41.2 & 50.0 \\
    BERT & 3.3 & 8.8 & 15.6 & 17.6 \\
    \bottomrule
    \end{tabular}
    \vspace{-1mm}
    \caption{
    Error rate %
    on natural and nonce evaluation stimuli, stratified by whether the subject--verb pair was seen (frequency $\geq$ 1) or unseen (frequency = 0) during pre-training. 
    Heuristics (argmaxes) show performance for a item-specific learner that memorizes probabilities of specific verbs (V) or subject--verb (SV) pairs. 
    BERT's performance degrades on unseen pairs, but is significantly better than these heuristics. 
    }
    \label{tab:novel_sv}
    \vspace{-2mm}
\end{table}
\endgroup
\subsection{Unseen Subject--Verb Pairs}\label{subsec:label_unseen_sv_pairs}
Table \ref{tab:novel_sv} stratifies error rate by seen and unseen subject--verb pairs. 
Compared with subject--verb pairs seen at least once during training, error rates on unseen subject--verb pairs are 5\% higher on natural sentences and 2\% higher on nonce sentences. 
This degradation, however, is minimal compared with what we might expect from a naive item-specific learner (H2), represented by the heuristic baselines in Table \ref{tab:novel_sv}. 
These results thus suggest that BERT reasons over representations that abstract to some degree over individual words, though it does not meet the definition of a fully-abstract symbolic learner (H1), which would have no degredation in performance. %

\subsection{Frequency of the Target Form}\label{subsec:raw_frequency}
To further examine the effect of frequency, we draw inspiration from the human language processing literature. 
One of the most widely-observed phenomena in such research is that high-frequency words are learned better (\citealt{ubiquitous}):

\begin{reh}
High-frequency forms reduce errors in contexts where they are the target.
\end{reh}

\cref{fig:sv_and_v_freq_fig} stratifies error rate by the training frequency of (1) subject--verb pairs and (2) verbs (independent of subject). 
On both natural and nonce stimuli, error rate decreases for more-frequent subject--verb pairs and more-frequent verbs, consistent with the Reduce Error hypothesis.   

\pgfplotsset{width=4.7cm, height=4.3cm}
\begin{figure}[t]
\small 
\begin{tikzpicture}
\begin{groupplot}[
       group style={
       group name=plot,
       horizontal sep=3pt,
       vertical sep=0pt,
       group size=2 by 1},]
  \nextgroupplot[
    ybar,
    title = {Natural Sentences},
    title style={xshift=0mm, yshift=-2mm, font=\small},
    symbolic x coords={0,1--10,11--100,101--1000,>1000},
    xtick={0,1--10,11--100,101--1000,>1000},
    x tick label style={rotate=45, anchor=north east, inner sep=0mm},
    xtick pos=bottom,
    enlarge x limits=0.2,
    xlabel={Frequency of subject--verb pair in the training set},
    xlabel style={xshift=15.5mm, yshift=0mm, font=\small},
    ymin=0,
    ymax=21,
    ytick={0, 5, 10, 15, 20},
    ytick pos=left,
    ylabel={Error rate (\%)},
    ylabel style={xshift=0mm, yshift=-1.4mm, font=\small},
    every axis plot/.append style={
        bar shift=0pt,
        draw=black, 
    },
    ]
    \addplot[fill=bluegold5!80,error bars/.cd, y dir=both, y explicit] 
        coordinates { (0,8.9) += (0,5.7) -= (0,6.6)};
    \addplot[fill=bluegold5!80,error bars/.cd, y dir=both, y explicit] 
        coordinates { (1--10,8.7) += (0,2.9) -= (0,3.3)};
    \addplot[fill=bluegold5!80,error bars/.cd, y dir=both, y explicit] 
        coordinates { (11--100,4.5) += (0,0.8) -= (0,0.8)};
    \addplot[fill=bluegold5!80,error bars/.cd, y dir=both, y explicit] 
        coordinates { (101--1000,4.0) += (0,0.4) -= (0,0.4)};
    \addplot[fill=bluegold5!80,error bars/.cd, y dir=both, y explicit] 
        coordinates { (>1000,2.3) += (0,0.3) -= (0,0.2)};
  \nextgroupplot[
    ybar,
    title = {Nonce Sentences},
    title style={xshift=0mm, yshift=-2mm, font=\small},
    symbolic x coords={0,1--10,11--100,101--1000,>1000},
    xtick={0,1--10,11--100,101--1000,>1000},
    x tick label style={rotate=45, anchor=north east, inner sep=0mm},
    xtick pos=bottom,
    enlarge x limits=0.2,
    ymin=0,
    ymax=21,
    ytick={0, 5, 10, 15, 20},
    yticklabels={,,},
    ytick pos=left,
    ylabel style={xshift=0mm, yshift=-1.4mm, font=\small},
    every axis plot/.append style={
        bar shift=0pt,
        draw=black, 
    }
    ]
    \addplot[fill=bluegold5!80,error bars/.cd, y dir=both, y explicit] 
        coordinates { (0,17.6) += (0,0.1) -= (0,0.1)};
    \addplot[fill=bluegold5!80,error bars/.cd, y dir=both, y explicit] 
        coordinates { (1--10,15.4) += (0,0.1) -= (0,0.1)};
    \addplot[fill=bluegold5!80,error bars/.cd, y dir=both, y explicit] 
        coordinates { (11--100,16) += (0,0.1) -= (0,0.1)};
    \addplot[fill=bluegold5!80,error bars/.cd, y dir=both, y explicit] 
        coordinates { (101--1000,15.5) += (0,0.1) -= (0,0.1)};
    \addplot[fill=bluegold5!80,error bars/.cd, y dir=both, y explicit] 
        coordinates { (>1000,13.4) += (0,0.4) -= (0,0.5)};
\end{groupplot}
\end{tikzpicture}
\\

\begin{tikzpicture}
\begin{groupplot}[
      group style={
      group name=plot,
      horizontal sep=3pt,
      vertical sep=0pt,
      group size=2 by 1},]
  \nextgroupplot[
    ybar,
    title = {Natural Sentences},
    title style={xshift=0mm, yshift=-2mm, font=\small},
    symbolic x coords={<50k,50k--100k,100k--200k,200k--400k,400k--800k,>800k},
    xtick={<50k,50k--100k,100k--200k,200k--400k,400k--800k,>800k},
    x tick label style={rotate=45, anchor=north east, inner sep=0mm},
    xtick pos=bottom,
    enlarge x limits=0.13,
    xlabel={Frequency of target verb in the training set},
    xlabel style={xshift=17mm, yshift=0mm, font=\small},
    ymin=0,
    ymax=42,
    ytick={0, 10, 20, 30, 40},
    ytick pos=left,
    ylabel={Error rate (\%)},
    ylabel style={xshift=0mm, yshift=-1.4mm, font=\small},
    every axis plot/.append style={
        bar shift=0pt,
        draw=black, 
    }
    ]
    \addplot[fill=amethyst!80,error bars/.cd, y dir=both, y explicit] 
        coordinates { (<50k,14.3) += (0,1.4) -= (0,1.44)};
    \addplot[fill=amethyst!80,error bars/.cd, y dir=both, y explicit] 
        coordinates { (50k--100k,12.5) += (0,0.85) -= (0,0.83)};
    \addplot[fill=amethyst!80,error bars/.cd, y dir=both, y explicit] 
        coordinates { (100k--200k,8.5) += (0,0.63) -= (0,0.53)};
    \addplot[fill=amethyst!80,error bars/.cd, y dir=both, y explicit] 
        coordinates { (200k--400k,4.6) += (0,0.78) -= (0,0.84)};
    \addplot[fill=amethyst!80,error bars/.cd, y dir=both, y explicit] 
        coordinates { (400k--800k,0.5) += (0,0.16) -= (0,0.15)};
    \addplot[fill=amethyst!80,error bars/.cd, y dir=both, y explicit] 
        coordinates { (>800k,1.9) += (0,0.51) -= (0,0.46)};
    
  \nextgroupplot[
    ybar,
    title = {Nonce Sentences},
    title style={xshift=0mm, yshift=-2mm, font=\small},
    symbolic x coords={<25k,25k--50k,50k--100k,100k--200k,200k--400k,400k--800k,>800k},
    xtick={<25k,25k--50k,50k--100k,100k--200k,200k--400k,400k--800k,>800k},
    x tick label style={rotate=45, anchor=north east, inner sep=0mm},
    xtick pos=bottom,
    enlarge x limits=0.13,
    ymin=0,
    ymax=42,
    ytick={0, 10, 20, 30, 40},
    yticklabels={,,},
    ytick pos=left,
    ylabel style={xshift=0mm, yshift=-1.4mm, font=\small},
    every axis plot/.append style={
        bar shift=0pt,
        draw=black, 
    }
    ]
    \addplot[fill=amethyst!80,error bars/.cd, y dir=both, y explicit] 
        coordinates { (<25k,33.5) += (0,0.1) -= (0,0.1)};
    \addplot[fill=amethyst!80,error bars/.cd, y dir=both, y explicit] 
        coordinates { (25k--50k,21) += (0,0.1) -= (0,0.1)};
    \addplot[fill=amethyst!80,error bars/.cd, y dir=both, y explicit] 
        coordinates { (50k--100k,18.5) += (0,0.1) -= (0,0.1)};
    \addplot[fill=amethyst!80,error bars/.cd, y dir=both, y explicit] 
        coordinates { (100k--200k,14.7) += (0,0.1) -= (0,0.1)};
    \addplot[fill=amethyst!80,error bars/.cd, y dir=both, y explicit] 
        coordinates { (200k--400k,12.6) += (0,0.1) -= (0,0.1)};
    \addplot[fill=amethyst!80,error bars/.cd, y dir=both, y explicit] 
        coordinates { (400k--800k,9.7) += (0,0.1) -= (0,0.1)};
    \addplot[fill=amethyst!80,error bars/.cd, y dir=both, y explicit] 
        coordinates { (>800k,9) += (0,0.1) -= (0,0.1)};
\end{groupplot}
\end{tikzpicture}
\caption{
Error rate stratified by how often subject--verb pairs appeared in the same sentence in BERT's training set (top) and how often verbs appeared in the training set (bottom).
Error rate was lower for subject--verb pairs and verbs that were more frequent.
}
\label{fig:sv_and_v_freq_fig}
\end{figure}
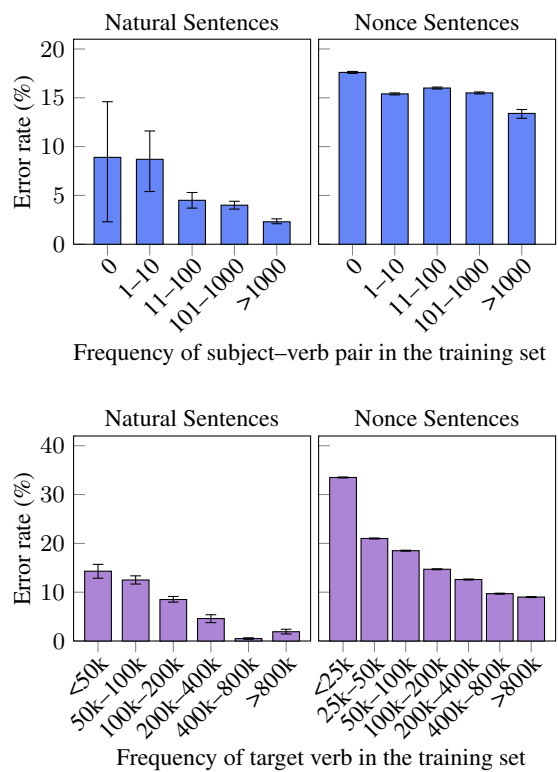

\subsection{Frequency of the Competing Form}\label{subsec:ratio_frequency}
Although seeing a verb more often in training often reduces errors when that verb is the target, when high-frequency forms are not the target, they can act as distractors and reduce accuracy:

\begin{ceh}
High-frequency forms cause errors when a competing, lower-frequency form is the target \cite{ubiquitous}.
\end{ceh}

\pgfplotsset{width=4.9cm, height=4.6cm}
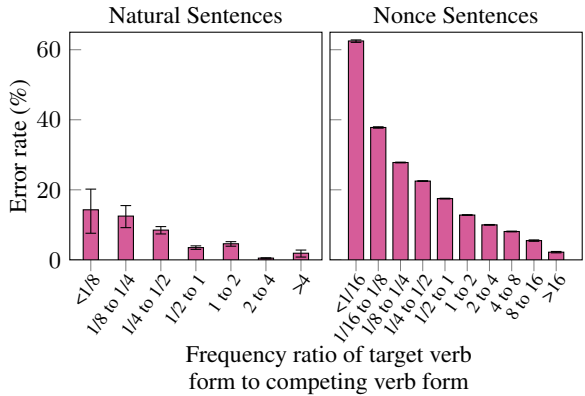
\begin{figure}[t]
\small 
\begin{tikzpicture}
\begin{groupplot}[
      group style={
      group name=plot,
      horizontal sep=3pt,
      vertical sep=0pt,
      group size=2 by 1},]
  \nextgroupplot[
    ybar,
    title = {Natural Sentences},
    title style={xshift=0mm, yshift=-2mm, font=\small},
    symbolic x coords={<1/8,1/8 to 1/4,1/4 to 1/2,1/2 to 1,1 to 2,2 to 4,>4},
    xtick={<1/8,1/8 to 1/4,1/4 to 1/2,1/2 to 1,1 to 2,2 to 4,>4},
    x tick label style={rotate=55, anchor=north east, inner sep=0mm, font=\scriptsize},
    xtick pos=bottom,
    enlarge x limits=0.1,
    xlabel={Frequency ratio of target verb\\ form to competing verb form},
    xlabel style={xshift=17.5mm, yshift=0mm, font=\small, align=center},
    bar width=2mm,
    ymin=0,
    ymax=65,
    ytick={0, 20, 40, 60},
    ytick pos=left,
    ylabel={Error rate (\%)},
    ylabel style={xshift=0mm, yshift=-1.4mm, font=\small},
    every axis plot/.append style={
        bar shift=0pt,
        draw=black, 
    }
    ]
    \addplot[fill=uclagoldfake!80,error bars/.cd, y dir=both, y explicit] 
        coordinates { (<1/8,14.3) += (0,5.9) -= (0,6.7)};
    \addplot[fill=uclagoldfake!80,error bars/.cd, y dir=both, y explicit] 
        coordinates { (1/8 to 1/4,12.5) += (0,3) -= (0,3.3)};
    \addplot[fill=uclagoldfake!80,error bars/.cd, y dir=both, y explicit] 
        coordinates { (1/4 to 1/2,8.5) += (0,1) -= (0,1.1)};
    \addplot[fill=uclagoldfake!80,error bars/.cd, y dir=both, y explicit] 
        coordinates { (1/2 to 1,3.5) += (0,0.5) -= (0,0.5)};
    \addplot[fill=uclagoldfake!80,error bars/.cd, y dir=both, y explicit] 
        coordinates { (1 to 2,4.6) += (0,0.6) -= (0,0.7)};
    \addplot[fill=uclagoldfake!80,error bars/.cd, y dir=both, y explicit] 
        coordinates { (2 to 4,0.5) += (0,0.1) -= (0,0.2)};
    \addplot[fill=uclagoldfake!80,error bars/.cd, y dir=both, y explicit] 
        coordinates { (>4,1.9) += (0,0.9) -= (0,1.1)};
    
  \nextgroupplot[
    ybar,
    title = {Nonce Sentences},
    title style={xshift=0mm, yshift=-2mm, font=\small},
    symbolic x coords={<1/16,1/16 to 1/8,1/8 to 1/4,1/4 to 1/2,1/2 to 1,1 to 2,2 to 4,4 to 8,8 to 16,>16},
    xtick={<1/16,1/16 to 1/8,1/8 to 1/4,1/4 to 1/2,1/2 to 1,1 to 2,2 to 4,4 to 8,8 to 16,>16},
    x tick label style={rotate=55, anchor=north east, inner sep=0mm, font=\scriptsize},
    xtick pos=bottom,
    enlarge x limits=0.13,
    bar width=2mm,
    ymin=0,
    ymax=65,
    ytick={0, 20, 40, 60},
    yticklabels={,,},
    ytick pos=left,
    ylabel style={xshift=0mm, yshift=-1.4mm, font=\small},
    every axis plot/.append style={
        bar shift=0pt,
        draw=black, 
    }
    ]
    \addplot[fill=uclagoldfake!80,error bars/.cd, y dir=both, y explicit] 
        coordinates { (<1/16,62.5) += (0,0.3) -= (0,0.4)};
    \addplot[fill=uclagoldfake!80,error bars/.cd, y dir=both, y explicit] 
        coordinates { (1/16 to 1/8,37.8) += (0,0.2) -= (0,0.2)};
    \addplot[fill=uclagoldfake!80,error bars/.cd, y dir=both, y explicit] 
        coordinates { (1/8 to 1/4,27.8) += (0,0.1) -= (0,0.1)};
    \addplot[fill=uclagoldfake!80,error bars/.cd, y dir=both, y explicit] 
        coordinates { (1/4 to 1/2,22.5) += (0,0.1) -= (0,0.1)};
    \addplot[fill=uclagoldfake!80,error bars/.cd, y dir=both, y explicit] 
        coordinates { (1/2 to 1,17.5) += (0,0.1) -= (0,0.1)};
    \addplot[fill=uclagoldfake!80,error bars/.cd, y dir=both, y explicit] 
        coordinates { (1 to 2,12.8) += (0,0.1) -= (0,0.1)};
    \addplot[fill=uclagoldfake!80,error bars/.cd, y dir=both, y explicit] 
        coordinates { (2 to 4,10.0) += (0,0.1) -= (0,0.1)};
    \addplot[fill=uclagoldfake!80,error bars/.cd, y dir=both, y explicit] 
        coordinates { (4 to 8,8.1) += (0,0.1) -= (0,0.1)};
    \addplot[fill=uclagoldfake!80,error bars/.cd, y dir=both, y explicit] 
        coordinates { (8 to 16,5.5) += (0,0.2) -= (0,0.2)};
    \addplot[fill=uclagoldfake!80,error bars/.cd, y dir=both, y explicit] 
        coordinates { (>16,2.2) += (0,0.2) -= (0,0.2)};
\end{groupplot}
\end{tikzpicture}
\vspace{-4.5mm}
\caption{
Error rate stratified by the ratio between the frequency of the target verb form versus the competing verb form.
BERT's error rate was higher when the competing verb form occurred more frequently in the training set than the target verb form.
}
\vspace{-1.5mm}
\label{fig:freq_asymmetry}
\end{figure}

\noindent Is BERT's error rate similarly affected by distractor frequency effects? 
For instance, the word \textit{``combat,''} which is not only the plural form of the verb \textit{``combat''} but also a fairly frequent noun, appears $102\times$ more often in the training set than \textit{``combats.''} 
If word frequency influences BERT's predictions, then such asymmetries may cause a high error rate when the target form is \textit{``combats.''}

As \cref{fig:freq_asymmetry} shows, error rate is lower when the target form is more frequent relative to the competing form. 
For nonce sentences, for example, error rate was only 2.2\% when the target form was 16 times or more as frequent than the competing form, compared with 62.5\% when the competing form was 16 times or more frequent than the target form. %

\subsection{Takeaways}

The above exploratory analyses suggest that BERT is influenced by both the absolute frequency of the target form (Reduce Error Hypothesis), as well as the frequency of the target relative to the competing form (Cause Error Hypothesis). 
Although these results are strong correlational evidence, absolute and relative frequency are highly correlated with one another
(i.e., as the absolute frequency of a word increases, so does its frequency relative to other words).\footnote{This correlation is not only intuitive but also empirical---see \cref{fig:abs_vs_relative} in \cref{appendix:abs_vs_relative}.} %
Thus, more controlled studies are needed to establish which effects have a causal effect on BERT's rule-learning.

\section{Confirmatory Analysis: Manipulating the Training Data}\label{sec:manipulation}
To better understand the above trends, we design a set of experiments in which we manipulate one variable (absolute or relative frequency of a verb form in pre-training) while holding the other fixed.

\subsection{Experimental Setup}
We select 60 verbs of interest (VOIs) and manipulate their training set frequencies.
We choose the VOIs by taking 60 transitive verbs for which both singular and plural forms of each verb occur at least $10^4$ times in the corpus (the full list is shown in \cref{appendix:voi}).
We remove all sentences containing these VOIs from the training set, and, based on the experiment, add them back in such that VOIs appear at a specified (absolute or relative) frequency.
We evaluate the model's performance on these VOIs by using both a natural dataset of approximately 100 examples per VOI, as well as by inserting the VOIs into the nonce (noun + sentential context) constructions from \cref{subsubsec:nonce_stimuli}. 

We note that the exact size of the training set in our manipulations changes depending on how many sentences containing VOIs are added in (e.g., models which see 10,000 examples per VOI see more total training examples than models that see only 1 example per VOI). 
The difference in absolute terms, however, is small (less than 1\% of the total training set). 
Thus, we consider it unlikely that any observed difference in performance is due to a difference in the total size of the training corpus.\footnote{As one measure, masked language model accuracy (on the same dev set) was 59.96\% for a model with VOIs appearing at frequency 10,000, versus 59.91\% for a model with VOIs appearing at frequency 1.}

\subsection{Absolute Frequency of Verb Form}\label{subsec:mani_absolute_freq}
We first examine how the absolute frequency of a verb form affects the model's number agreement ability on that form.
For each of nine frequencies $n =$ 1, 3, 10, 30, 100, 300, 1,000, 3,000, and 10,000, we train a new BERT model that sees the verbs $n$ times each during training. 
To isolate the effect of absolute frequency, we fix the relative frequency to be balanced---for each VOI, $n$ instances are singular and $n$ are plural.

\begin{figure}[t]
\centering
\begin{tikzpicture}
    \pgfplotsset{small,samples=10}
    \begin{groupplot}[
        group style = {group size = 1 by 1}, 
        width = 5.2cm, 
        height = 4cm]
        \nextgroupplot[
            legend style={at={(1.03, 0.84)}, anchor=west},
            xlabel={Frequency of VOI in training set},
            ylabel={Error rate (\%)},
            xmode=log,
            xmin=0.8, xmax=15000,
            ymin=-5, ymax=55,
            xtick={1, 10, 100, 1000, 10000},
            ytick={0, 10, 20, 30, 40, 50},
            ymajorgrids=true,
            xmajorgrids=true,
            grid style=dashed,
            x label style={at={(axis description cs:0.5,-0.15)},anchor=north},
            y label style={at={(axis description cs:-0.1,0.5)},anchor=south},
            xtick pos=bottom,
            ytick pos=left,
            ]
            \addplot[
                color=blue,
                mark=diamond,
                mark size=3pt,
                ]
                coordinates {
                (1,     50.1)
                (3,     48.85)
                (10,    45.1)
                (30,    30.7)
                (100,   23.5)
                (300,   22.2)
                (1000,  20.1)
                (3000,  16.7)
                (10000, 14.95)
                };
                \addlegendentry{Nonce}
            \addplot[
                color=violet,
                mark=triangle,
                mark size=3pt,
                ]
                coordinates {
                (1,     48.65)
                (3,     45.7)
                (10,    35.3)
                (30,    13.4)
                (100,   4.5)
                (300,   4.25)
                (1000,  3.5)
                (3000,  2.9)
                (10000, 2.6)
                };  
                \addlegendentry{Natural}
    \end{groupplot}
\end{tikzpicture}
\vspace{-1mm}
\caption{
Effect of absolute frequency of a verb of interest (VOI) when the ratio between singular and plural forms is held constant at 1:1.
The error rate for sixty VOI is shown for BERT models that have seen the sixty VOI at different frequencies in the pre-training dataset.
}
\vspace{-3mm}
\label{fig:mani_raw_freq}
\end{figure}
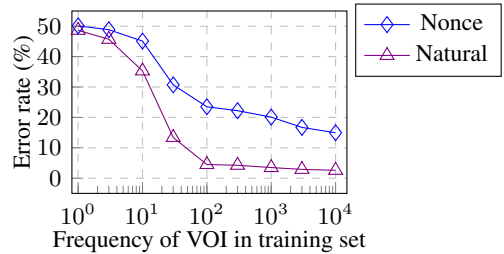

The results of this experiment are shown in \cref{fig:mani_raw_freq}. 
When the occurrences of a VOI are balanced between inflections (singular and plural), error rate decreases monotonically when target form is more frequent in training. 

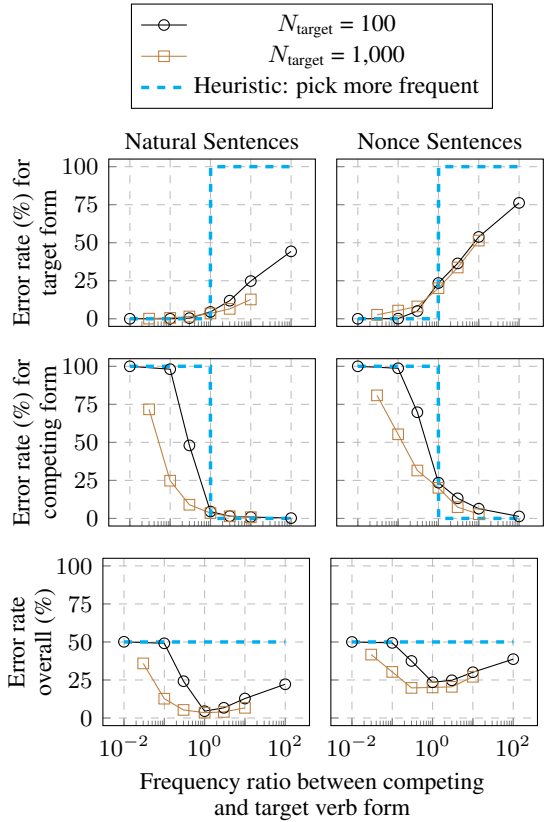
\begin{figure}[t]
\begin{centering}
\begin{tikzpicture}
    \pgfplotsset{footnotesize,samples=10}
    \begin{groupplot}[
        group style = {group size = 2 by 1, horizontal sep = 8pt}, 
        width = 4.3cm, 
        height = 3.8cm]
        \nextgroupplot[
            title = {Natural Sentences},
            legend style={at={(0.5,1.5)},anchor=north},
            xmode=log,
            xmin=0.003, xmax=400,
            ymin=-5, ymax=105,
            xtick={0.01, 0.1, 1, 10, 100},
            xticklabels={,,},
            ylabel={Error rate (\%) for \\target form},
            ylabel style={align=center},
            ytick={0, 25, 50, 75, 100},
            ymajorgrids=true,
            xmajorgrids=true,
            grid style=dashed,
            x label style={at={(axis description cs:0.5,-0.15)},anchor=north},
            y label style={at={(axis description cs:-0.18,0.5)},anchor=south},
            xtick pos=bottom,
            ytick pos=left,
            ]
            \addplot[
                color=black,
                mark=o,
                mark size=2pt,
                ]
                coordinates {
                (0.01,  0)
                (0.1,   0)
                (0.3,   0.4)
                (1.0,   4.5)
                (3.0,   11.9)
                (10.0,  24.8)
                (100.0, 44.3)
                };
            \addplot[
                color=brown,
                mark=square,
                mark size=2pt,
                ]
                coordinates {
                (0.03,  0.1)
                (0.1,   0.6)
                (0.3,   1.5)
                (1,     3.5)
                (3,     6.5)
                (10,    12.7)
                };
            \addplot[
                color=cyan,
                mark=square,
                no markers,
                dashed,
                line width=1.3pt,
                ]
                coordinates {
                (0.01,   0)
                (0.99,   0)
                (1.01,   100)
                (100,    100)
                };
        \nextgroupplot[
            title = {Nonce Sentences},
            legend style={at={(-0.12,1.3)},anchor=south},
            xmode=log,
            xmin=0.003, xmax=400,
            ymin=-5, ymax=105,
            xtick={0.01, 0.1, 1, 10, 100},
            xticklabels={,,},
            ytick={0, 25, 50, 75, 100},
            yticklabels={,,},
            ymajorgrids=true,
            xmajorgrids=true,
            grid style=dashed,
            x label style={at={(axis description cs:-0.12,-0.22)},anchor=north},
            y label style={at={(axis description cs:-0.125,0.5)},anchor=south},
            xtick pos=bottom,
            ytick pos=left,
            ]
            \addplot[
                color=black,
                mark=o,
                mark size=2pt,
                ]
                coordinates {
                (0.01,  0)
                (0.1,   0.1)
                (0.3,   5.1)
                (1.0,   23.5)
                (3.0,   36.4)
                (10.0,  53.8)
                (100.0, 76.2)
                };
                \addlegendentry{$N_{\textrm{target}}$ = 100}
            \addplot[
                color=brown,
                mark=square,
                mark size=2pt,
                ]
                coordinates {
                (0.03,  2.6)
                (0.1,   5.4)
                (0.3,   8.1)
                (1,     20.1)
                (3,     33.7)
                (10,    51.4)
                };
                \addlegendentry{$N_{\textrm{target}}$ = 1,000}
            \addplot[
                color=cyan,
                mark=square,
                no markers,
                dashed,
                line width=1.3pt,
                ]
                coordinates {
                (0.01,   0)
                (0.99,   0)
                (1.01,   100)
                (100,    100)
                };
                \addlegendentry{Heuristic: pick more frequent}
    \end{groupplot}
\end{tikzpicture}
\begin{tikzpicture}
    \pgfplotsset{footnotesize,samples=10}
    \begin{groupplot}[
        group style = {group size = 2 by 1, horizontal sep = 8pt}, 
        width = 4.3cm, 
        height = 3.8cm]
        \nextgroupplot[
            legend style={at={(0.5,1.5)},anchor=north},
            xmode=log,
            xmin=0.003, xmax=400,
            ymin=-5, ymax=105,
            xtick={0.01, 0.1, 1, 10, 100},
            xticklabels={,,},
            ylabel={Error rate (\%) for\\ competing form},
            ylabel style={align=center},
            ytick={0, 25, 50, 75, 100},
            ymajorgrids=true,
            xmajorgrids=true,
            grid style=dashed,
            x label style={at={(axis description cs:0.5,-0.15)},anchor=north},
            y label style={at={(axis description cs:-0.18,0.5)},anchor=south},
            xtick pos=bottom,
            ytick pos=left,
            ]
            \addplot[
                color=black,
                mark=o,
                mark size=2pt,
                ]
                coordinates {
                (0.01,  100)
                (0.1,   98.1)
                (0.3,   48.0)
                (1.0,   4.5)
                (3.0,   1.5)
                (10.0,  0.9)
                (100.0, 0.2)
                };
            \addplot[
                color=brown,
                mark=square,
                mark size=2pt,
                ]
                coordinates {
                (0.03,  71.7)
                (0.1,   24.8)
                (0.3,   9.0)
                (1,     3.5)
                (3,     1.4)
                (10,    0.8)
                };
            \addplot[
                color=cyan,
                mark=square,
                no markers,
                dashed,
                line width=1.3pt,
                ]
                coordinates {
                (0.01,  100)
                (0.99,   100)
                (1.01,   0)
                (100,    0)
                };
        \nextgroupplot[
            legend style={at={(-0.12,1.3)},anchor=south},
            xmode=log,
            xmin=0.003, xmax=400,
            ymin=-5, ymax=105,
            xtick={0.01, 0.1, 1, 10, 100},
            xticklabels={,,},
            ytick={0, 25, 50, 75, 100},
            yticklabels={,,},
            ymajorgrids=true,
            xmajorgrids=true,
            grid style=dashed,
            x label style={at={(axis description cs:-0.12,-0.22)},anchor=north},
            y label style={at={(axis description cs:-0.125,0.5)},anchor=south},
            xtick pos=bottom,
            ytick pos=left,
            ]
            \addplot[
                color=black,
                mark=o,
                mark size=2pt,
                ]
                coordinates {
                (0.01,  100)
                (0.1,   98.7)
                (0.3,   69.8)
                (1.0,   23.5)
                (3.0,   13.2)
                (10.0,  6.4)
                (100.0, 1.3)
                };
            \addplot[
                color=brown,
                mark=square,
                mark size=2pt,
                ]
                coordinates {
                (0.03,  80.9)
                (0.1,   55.3)
                (0.3,   31.5)
                (1,     20.1)
                (3,     7.4)
                (10,    2.8)
                };
            \addplot[
                color=cyan,
                mark=square,
                no markers,
                dashed,
                line width=1.3pt,
                ]
                coordinates {
                (0.01,  100)
                (0.99,   100)
                (1.01,   0)
                (100,    0)
                };
    \end{groupplot}
\end{tikzpicture}
\begin{tikzpicture}
    \pgfplotsset{footnotesize,samples=10}
    \begin{groupplot}[
        group style = {group size = 2 by 1, horizontal sep = 8pt}, 
        width = 4.3cm, 
        height = 3.8cm]
        \nextgroupplot[
            legend style={at={(0.5,1.5)},anchor=north},
            xmode=log,
            xmin=0.003, xmax=400,
            ymin=-5, ymax=105,
            xtick={0.01, 0.1, 1, 10, 100},
            xticklabels={$10^{-2}$, , $10^{0}$, ,$10^{2}$},
            ylabel={Error rate \\ overall (\%)},
            ylabel style={align=center},
            ytick={0, 25, 50, 75, 100},
            ymajorgrids=true,
            xmajorgrids=true,
            grid style=dashed,
            x label style={at={(axis description cs:0.5,-0.15)},anchor=north},
            y label style={at={(axis description cs:-0.18,0.5)},anchor=south},
            xtick pos=bottom,
            ytick pos=left,
            ]
            \addplot[
                color=black,
                mark=o,
                mark size=2pt,
                ]
                coordinates {
                (0.01,  50.1)
                (0.1,   49.1)
                (0.3,   24.2)
                (1.0,   4.5)
                (3.0,   6.7)
                (10.0,  12.8)
                (100.0, 22.2)
                };
            \addplot[
                color=brown,
                mark=square,
                mark size=2pt,
                ]
                coordinates {
                (0.03,  35.9)
                (0.1,   12.7)
                (0.3,   5.3)
                (1,     3.5)
                (3,     4.0)
                (10,    6.7)
                };
            \addplot[
                color=cyan,
                mark=square,
                no markers,
                dashed,
                line width=1.3pt,
                ]
                coordinates {
                (0.01,   50)
                (100,    50)
                };
        \nextgroupplot[
            legend style={at={(-0.12,1.3)},anchor=south},
            xmode=log,
            xmin=0.003, xmax=400,
            ymin=-5, ymax=105,
            xtick={0.01, 0.1, 1, 10, 100},
            xticklabels={$10^{-2}$, , $10^{0}$, ,$10^{2}$},
            xlabel={Frequency ratio between competing\\ and target verb form},
            xlabel style={align=center},
            ytick={0, 25, 50, 75, 100},
            yticklabels={,,},
            ymajorgrids=true,
            xmajorgrids=true,
            grid style=dashed,
            x label style={at={(axis description cs:-0.09,-0.22)},anchor=north},
            y label style={at={(axis description cs:-0.125,0.5)},anchor=south},
            xtick pos=bottom,
            ytick pos=left,
            ]
            \addplot[
                color=black,
                mark=o,
                mark size=2pt,
                ]
                coordinates {
                (0.01,  50.0)
                (0.1,   49.4)
                (0.3,   37.5)
                (1.0,   23.5)
                (3.0,   24.8)
                (10.0,  30.1)
                (100.0, 38.7)
                };
            \addplot[
                color=brown,
                mark=square,
                mark size=2pt,
                ]
                coordinates {
                (0.03,  41.7)
                (0.1,   30.3)
                (0.3,   19.8)
                (1,     20.1)
                (3,     20.5)
                (10,    27.1)
                };
            \addplot[
                color=cyan,
                mark=square,
                no markers,
                dashed,
                line width=1.3pt,
                ]
                coordinates {
                (0.01,   50)
                (100,    50)
                };
    \end{groupplot}
\end{tikzpicture}
\vspace{-1.5mm}
\caption{
When the absolute frequency of the target verb form is held constant, increasing the relative frequency of the competing verb form increases error for the target form and decreases error for the competing form. 
This behavior is in the same direction as a heuristic that, at inference time, picks the more frequent verb.
}
\vspace{-3mm}
\label{fig:mani_ratio_freq}
\end{centering}
\end{figure}
\subsection{Relative Frequency of Verb Form}\label{subsec:mani_relative_freq}
We next analyze whether the frequency ratio between a target verb form $v$ and its competing form $v'$ affects the model's ability to produce $v$ in context.
To balance how often the target $v$ is singular vs.\ plural, we use the following procedure.
We randomly split our 60 VOI into two groups of 30 verbs each, which we denote as $\calS$ and $\calP$. 
In each experiment, we set the frequency of the singular verbs in $\calS$ to $N_{\textrm{vary}}$, while holding the frequency of the plural forms of the verbs in $\calS$ constant at $N_{\textrm{constant}}$. 
Likewise, we set the frequency of the plural verbs in $\calP$ to $N_{\textrm{vary}}$, and hold the frequency of the singular form of these verbs constant at $N_{\textrm{constant}}$. 
We run experiments with $N_{\textrm{vary}}=$ \{1, 10, 30, 100, 300, 1,000, 3,000, 10,000\}, and do this twice for $N_{\textrm{constant}}$ set to 100 and 1,000. 
As our evaluation stimuli are balanced such that both $v$ and $v'$ occur as the target in every template for every VOI, we are able to analyze the effect of the $v$:$v'$ frequency ratio---holding the absolute frequency of $v$ fixed---for $v$:$v'$ ranging from 1:100 to 100:1. 

\cref{fig:mani_ratio_freq} shows the results.
When the competing form occurs more frequently (with respect to the target form), error rate increases for the target form and decreases for the competing form.

\section{Disentangling Sources of Error}\label{sec:probe}

\subsection{Setup}\label{subsec:mani_logic}

Our goal is to characterize BERT's rule-learning behavior in terms of the three hypotheses H1--H3 described in \cref{sec:logic}. 
The frequency effects observed in \cref{sec:manipulation} rule out H1 (Idealized Symbolic Learner). 
However, BERT's generalization to unseen noun-verb pairs (\cref{subsec:label_unseen_sv_pairs}) is too good to be explained by H2 (Item-Specific Learner). 
Hence, the hybrid H3 (Symbolic Learner with Noisy Observations) seems like the most plausible candidate.

H3 is not simply a catch-all compromise between rule-based and item-specific learners---H3 makes specific predictions about the nature of the errors BERT will make. 
Under H3, BERT represents the SVA rule and the concept of agreement features, and follows the rule as long as it identifies the number of the subject and verb correctly. 
Thus, H3 predicts that observed errors are due to failures to identify the number of either the subject or verb.  

Given such a model, we might observe frequency effects because training frequency influences the model's ability to predict the agreement feature for a given verb form.
That is, we might observe a trend like the following: if a verb $v$ occurs in fewer than some $n$ training examples, BERT mispredicts the agreement feature (e.g., predicting $v$ to be singular when $v$ is plural); if $v$ occurs more than $n$ times, BERT correctly predicts $v$'s agreement feature and correctly produces $v$ in context.
In this scenario, we expect that SVA errors will correlate with frequency, but the frequency of these errors should not exceed the error rate in predicting agreement features.

\subsection{Predicting Agreement Feature}\label{subsec:mani_probe}
 
To test whether the above predicted pattern holds, we use two probing classifiers \cite[see][on probing]{veldhoen2016diagnostic,ettinger-etal-2016-probing} which we describe below.

\paragraph{Subject agreement feature probe.}
Our first probe evaluates whether, given a sentence with the verb masked, the embedding at the masked position contains information as to whether a singular or plural verb is required.
This setup actually evaluates two subtasks: identifying the subject of the verb (i.e., parsing the sentence) and predicting the agreement feature of the identified subject. 
For simplicity, however, we use a single probing classifier because our interpretation does not hinge on differentiating these subtasks. 
We use our sentential templates for this experiment, for which the only cue for the number of the subject (and hence the verb) is the subject itself.
Hence, if the embedding at the masked position can be used to predict number, it follows that both the subject has been identified correctly and that the agreement feature of the subject was identified correctly.

We feed our nonce sentences from \cref{subsubsec:nonce_stimuli} with the verb masked into the model, retrieve the final hidden state representation of the masked token, and train an MLP to classify the desired verb form (singular or plural). 
We train the probe using cross-validation, using sentences constructed from 150 subjects $\times$ 50 sentential for training, and the remainder for evaluation.
Subjects and sentential contexts in evaluation sentences are not seen by the probe during training.

\paragraph{Verb agreement feature probe.}
Our second probe predicts the number of a verb from its contextual word embedding.
If a probe can predict the number of a verb given its contextual word embedding, we can conclude that the model represents the agreement feature of that verb form and thus its predictions about SVA can, in principle, depend on the agreement feature. 
We obtain contextual embeddings of verbs by inserting them into the nonce sentential contexts, with the noun masked so that there are no external clues aside from the form of the verb that indicate whether the verb is singular or plural.
We then train an MLP to, given the embedding, classify the verb as singular or plural.
We use the 331 verbs from our nonce stimuli that were not VOI ($\times$56 sentential contexts per verb) to train the probe and the 60 VOI ($\times$56 sentential contexts per verb) to evaluate it.

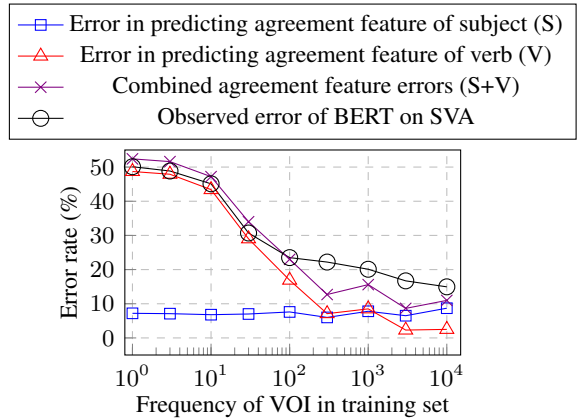
\begin{figure}[t]
\centering
\begin{tikzpicture}
    \pgfplotsset{small,samples=10}
    \begin{groupplot}[
        group style = {group size = 1 by 1}, 
        width = 6.0cm, 
        height = 4.3cm]
        \nextgroupplot[
            legend style={at={(0.5, 1.05)}, anchor=south},
            xlabel={Frequency of VOI in training set},
            ylabel={Error rate (\%)},
            xmode=log,
            xmin=0.8, xmax=15000,
            ymin=-5, ymax=55,
            xtick={1, 10, 100, 1000, 10000},
            ytick={0, 10, 20, 30, 40, 50},
            ymajorgrids=true,
            xmajorgrids=true,
            grid style=dashed,
            x label style={at={(axis description cs:0.5,-0.15)},anchor=north},
            y label style={at={(axis description cs:-0.1,0.5)},anchor=south},
            xtick pos=bottom,
            ytick pos=left,
            ]
            \addplot[
                color=blue,
                mark=square,
                mark size=2pt,
                ]
                coordinates {
                (1,     7.2)
                (3,     7.1)
                (10,    6.8)
                (30,    7)
                (100,   7.6)
                (300,   6)
                (1000,  7.8)
                (3000,  6.5)
                (10000, 8.7)
                };
                \addlegendentry{Error in predicting agreement feature of subject (S)}
            \addplot[
                color=red,
                mark=triangle,
                mark size=3pt,
                ]
                coordinates {
                (1,     48.7)
                (3,     47.9)
                (10,    43.4)
                (30,    29)
                (100,   16.9)
                (300,   7.1)
                (1000,  8.5)
                (3000,  2.3)
                (10000, 2.5)
                };
                \addlegendentry{Error in predicting agreement feature of verb (V)}
            \addplot[
                color=violet,
                mark=x,
                mark size=3pt,
                ]
                coordinates {
                (1,     52.4)
                (3,     51.6)
                (10,    47.2)
                (30,    34)
                (100,   23)
                (300,   12.7)
                (1000,  15.6)
                (3000,  8.65)
                (10000, 10.98)
                };
                \addlegendentry{Combined agreement feature errors (S+V)}
            \addplot[
                color=black,
                mark=o,
                mark size=3pt,
                ]
                coordinates {
                (1,     50.1)
                (3,     48.85)
                (10,    45.1)
                (30,    30.7)
                (100,   23.5)
                (300,   22.2)
                (1000,  20.1)
                (3000,  16.7)
                (10000, 14.95)
                };
                \addlegendentry{Observed error of BERT on SVA}
    \end{groupplot}
\end{tikzpicture}
\vspace{-1.5mm}
\caption{
Errors in either identifying the number agreement feature of the subject or identifying the number agreement feature of the verb comprised a large portion of the observed SVA error.
}
\label{fig:mani_probe}
\end{figure}
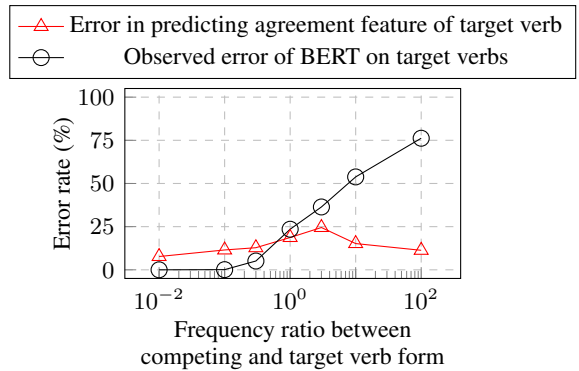
\begin{figure}[t]
\centering
\begin{tikzpicture}
    \pgfplotsset{small,samples=10}
    \begin{groupplot}[
        group style = {group size = 1 by 1}, 
        width = 6.0cm, 
        height = 4.1cm]
        \nextgroupplot[
            legend style={at={(0.5, 1.05)}, anchor=south},
            xlabel={Frequency ratio between \\competing and target verb form},
            xlabel style={align=center},
            ylabel={Error rate (\%)},
            ylabel style={align=center},
            ytick={0, 25, 50, 75, 100},
            xmode=log,
            xmin=0.003, xmax=400,
            ymin=-5, ymax=105,
            xtick={0.01, 0.1, 1, 10, 100},
            xticklabels={$10^{-2}$, , $10^{0}$, ,$10^{2}$},
            ymajorgrids=true,
            xmajorgrids=true,
            grid style=dashed,
            x label style={at={(axis description cs:0.5,-0.17)},anchor=north},
            y label style={at={(axis description cs:-0.12,0.5)},anchor=south},
            xtick pos=bottom,
            ytick pos=left,
            ]
            \addplot[
                color=red,
                mark=triangle,
                mark size=3pt,
                ]
                coordinates {
                (0.01,  7.7)
                (0.1,   11.5)
                (0.3,   12.8)
                (1,   18.7)
                (3,   24.5)
                (10,  15.2)
                (100,   11.3)
                };
                \addlegendentry{Error in predicting agreement feature of target verb}
            \addplot[
                color=black,
                mark=o,
                mark size=3pt,
                ]
                coordinates {
                (0.01,  0)
                (0.1,   0.1)
                (0.3,   5.1)
                (1.0,   23.5)
                (3.0,   36.4)
                (10.0,  53.8)
                (100.0, 76.2)
                };
                \addlegendentry{Observed error of BERT on target verbs}
    \end{groupplot}
\end{tikzpicture}
\vspace{-2mm}
\caption{
    The error rate of a probe that predicts the agreement feature of a verb (red triangle) is not correlated with frequency of competing verb forms. 
    Moreover, the error rate of this probe does not correlate with the observed error of BERT on the target verb, which is highly affected by frequency of competing verb forms.
}
\vspace{-3mm}
\label{fig:probe_ratio_new}
\end{figure}
\paragraph{Results.}
We evaluate these two probes as a function of both absolute frequency (using models from \cref{subsec:mani_absolute_freq}) and relative frequency (using models from \cref{subsec:mani_relative_freq}). Results are shown in \cref{fig:mani_probe} and \cref{fig:probe_ratio_new}, respectively.

For absolute frequency (\cref{fig:mani_probe}), we see that the accuracy of the verb agreement feature probe is highly dependent on absolute frequency of the VOI. The probe has lower error for models that saw the verb form more often, implying that seeing forms more frequently in training led to embeddings of those forms that better encode the number agreement feature. In constrast, the accuracy of the subject agreement feature probe is constant, which is expected because identifying the number feature of a subject should not be affected by absolute frequency of VOI.
Notably, the combined error rate of our two probes falls close to the model's observed overall error rate on the SVA task, as predicted by our ``Symbolic Reasoner with Noisy Observations'' hypothesis (H3).

For relative frequency (\cref{fig:probe_ratio_new}), on the other hand, we see no clear increase or decrease in the accuracy of predicting the agreement feature for a target verb form $v$ in response to changes in frequency of the competing verb form $v'$. 
In other words, when one verb is much more frequent than the other, BERT produces the more common verb form despite having access (in principle) to the information (rule + agreement features) that would allow it to infer the correct form. Such behavior is not explicitly accounted for by the ``noisy observations'' in H3, and thus appears more as evidence of item-specific learning (in line with H2).

\section{Discussion}

The goal of this work is to determine whether BERT performs SVA by implicitly representing rules defined over abstract agreement features and characterize the training conditions under which such representations emerge.
We differentiate between the representation of the rule (\textit{``if x then y''}) and that of observations (containing the correct agreement features).
We draw two conclusions, which suggest a mix of systematic rule-like generalization and unsystematic item-specific inferences. 

\paragraph{BERT appears to represent the correct rule, but fails to predict agreement features for low-frequency verb forms.}
Although the error rate decreases as a function of frequency of target verb $v$ (\cref{subsec:mani_absolute_freq}), BERT's ability to predict the agreement feature of $v$ (\cref{subsec:mani_probe}) follows the same trend. 
This observed behavior is thus consistent with a model that correctly represents the SVA rule (\cref{sec:logic}), but makes mistakes at inference time due to noise in the represented observations for low frequency verb forms (for example, producing \textit{``run''} in the context \textit{``the dog \underline{run}''} because \textit{``run''} is incorrectly encoded as singular), rather than due to a failure to represent the concept of singular altogether. 

\paragraph{BERT fails to apply the rule when doing so requires overcoming strong item-specific priors.} 
Similar to the absolute frequency trend, we see that BERT's error rate on SVA also decreases as a function of the frequency of the target verb $v$ relative to its competing form $v'$ (\cref{subsec:mani_relative_freq}). 
Unlike above, however, we see no effect of the frequency ratio $N_{v}:N_{v'}$ on BERT's ability to predict the agreement feature of $v$ when the frequency of $v$ is fixed (\cref{subsec:mani_probe}). 
These results suggest that BERT is heavily influenced by skewed training distributions, preferring to produce more common verb forms over forms consistent with the rule. 
Such behavior could either mean that, when $P(v) << P(v')$, (1) BERT represents the correct SVA rule but it is overridden in favor of the prior, or (2) BERT does not represent the rule at all.
Teasing apart these possibilities is a valuable direction for future work.

\paragraph{Open questions.} 
Our results on absolute frequency effects indicate that BERT does not infer agreement features until it sees 10--100 examples of a verb, even though it is possible, in principle, to infer agreement features from a single training example (e.g., \textit{``All of the dogs dax''} implies \textit{``dax''} is a plural verb form).
Future controlled studies could investigate how the sample efficiency of inferring agreement information depends on factors such as architecture (e.g., access to morphological signals), size of the model, and amount of training data. 
Analysis of such patterns would elucidate how models like BERT (and by extension, transformers and neural networks more generally) learn and generalize, enabling more principled development and deployment. 

\section{Conclusions}
We have studied whether BERT's performance on subject--verb agreement exhibits rule-governed behavior. 
We focus on frequency effects, pre-training multiple BERT instances in order to isolate how the model's predictions are affected by absolute and relative verb frequency.
Our results suggest that BERT's behavior is consistent with a system that correctly applies the SVA rule in general but struggles to overcome strong training priors and to estimate agreement features (singular vs.\ plural) on infrequent lexical items.

\section*{Acknowledgements}
We thank Ian Tenney, Ryan Cotterell, Clara Meister, Charles Lovering, and Slav Petrov for feedback on our manuscript.
We thank Christo Kirov, Kellie Webster, and Jacob Eisenstein for helpful discussions about the project.

\bibliography{custom}
\bibliographystyle{acl_natbib}

\clearpage
\appendix

\section{BERT Model}\label{appendix:bert}
To analyze the effect of word frequency on BERT's ability to follow SVA, we need to know the exact number of occurrences of each word in the dataset. 
The original BERT checkpoint \cite{devlin2018bert} uses both Wikipedia and BooksCorpus \cite{zhu2015aligning}, but BooksCorpus is no longer publicly available \cite{documentationdebt21}. 
So we train a replicated version of BERT on only wikipedia data.

Our version of BERT largely follows the procedure of the original, differing only in that we use dynamic masking and pre-train for 4 million updates at a learning rate of 3e-4.\footnote{The decision to use dynamic masking was made to the availability of code, rather theoretically or empirically motivated. We train for additional updates because the development loss did not converge at 1 million updates (the original number used in the paper).}
\cref{tab:bert_performance} shows the performance of our replicated version of BERT. 
\begingroup
\setlength{\tabcolsep}{2pt}
\begin{table}[ht]
    \centering
    \small
    \begin{tabular}{l | c c c c}
        \toprule
         & \multicolumn{4}{c}{Downstream Task} \\
        Pre-training data & MRPC & CoLA & MNLI & SST2 \\
        \midrule
        Wikipedia + BooksCorpus & 86.6 & 82.1 & 84.4 & 92.8 \\
        Wikipedia only & 86.2 & 78.7 & 84.3 & 92.2 \\
        \toprule
         & QQP & QNLI & RTE & STS-B \\
        \midrule
        Wikipedia + BooksCorpus & 91.2 & 91.6 & 68.5 & 89.3 \\
        Wikipedia only & 90.8 & 91.5 & 64.9 & 88.8 \\
        \bottomrule
    \end{tabular}
    \caption{
    Performance of our replicated BERT checkpoint, which is pre-trained on only Wikipedia data, compared with the original BERT checkpoint, which used both Wikipedia and BooksCorpus.
    }
    \label{tab:bert_performance}
\end{table}
\endgroup

\section{Nonce Stimuli Collection Details}\label{appendix:non_dataset_details}
This appendix section details our nonce stimuli collection process.
Our goal is to create a large set of evaluation stimuli in which we can analyze how properties of certain stimuli (e.g., subjects, verbs, and sentential contexts) affect the model's ability to perform number agreement. 
Therefore, we create a list of 200 nouns, 336 verbs, and 56 sentential contexts such that any given (noun, verb, sentential context) triplet where the noun is used as the subject yields a grammatically correct nonce sentence. 
By considering both singular and plural inflections for possible triplet, we analyze a dataset of 2 $\cdot$ 200 $\cdot$ 336 $\cdot$ 56 = 7,526,400 sentences.
To facilitate further use of our dataset, we make a plain-text version available at \url{http://anonymized}.

\subsection{Nouns}
To propose candidate nouns, we first ran a POS tagger \cite{tsai-etal-2019-small} over the pre-training dataset, and retrieved all nouns occurring at least 100 times. 
Then, we randomly sampled 200 nouns from this set of candidate nouns that were evenly distributed into four buckets of training set frequency (100--999, 1,000--9,999, 10,000--99,999, and 100,000+). 
All nouns were common nouns and were manually validated to have correct, unambiguous singular and plural inflections.

\subsection{Verbs}
To propose candidate verbs, we similarly retrieved all verbs that occurred at least 100 times in the training set. 
The masked-LM evaluation procedure for SVA requires that both the singular and plural inflections of the verb exist directly in the model's vocabulary, and so we filtered out verbs that did not match this criteria, leaving us with 379 candidate verbs. 

Unlike for nouns, we generally cannot indiscriminately swap out verbs in a sentence while maintaining grammatical correctness, since some verbs are exclusively transitive (used with an object) or intransitive (used without an object). 
In English, more verbs can be used transitively than intransitively, and so we decided to consider only transitive verbs.
We manually filtered out strictly intransitive verbs and ensured that each verb had correct, unambiguous singular and plural inflections, leaving us with 336 verbs that can be used transitively.

\subsubsection{Sentential Contexts}
Finally, we curated a list of sentential contexts (sentences with the subject and verb removed) that would maintain grammatical validity for both singular and plural forms of any given subject--verb pair from our list of nouns and list of transitive verbs. 
To get candidate sentential contexts, we randomly sampled 600 sentences from the \citet{linzen-etal-2016-assessing} dataset of Wikipedia sentences to be manually examined.
We kept only 56 of these 600 candidate sentential contexts, filtering out 544 for the following reasons: 
\begin{itemize}[leftmargin=*]
    \item 
    Sentential contexts that contained additional cues for number outside of the subject and verb inflection cannot form grammatical sentences for both singular and plural subject--verb pairs.
    For instance, the sentential context ``\textit{[SUBJECT], who \underline{thinks} roses are red, [VERB] ...}'' can only be used with singular subjects and verbs because of the modifying clause ``\textit{who thinks roses are red}''; and the sentential context ``\textit{[SUBJECT] in the park [VERB] ...}'' can only be used with plural subjects and verbs because there is no determiner for the subject.
    381 sentences like the above had such cues for number inflection and were removed.
    \item 
    64 sentential contexts contained verb usages that were hostile to swapping in most transitive verbs (e.g., in ``\textit{[SUBJECT] \underline{shows that} ...}'', ``\textit{shows}'' could not be replaced with most transitive verbs).
    \item 
    15 sentential contexts contained noun usages that were hostile to swapping in most nouns (e.g., in \textit{the fact that she likes him ...}, the subject \textit{fact} cannot be replaced with most nouns).
    \item 
    21 sentential contexts were ungrammatical or incomprehensible to our human annotator.
    \item 
    In 36 sentential contexts, the subject and verb parsed by \citet{linzen-etal-2016-assessing} was incorrect.
    \item 
    27 sentential contexts for which the original verb was used intransitively were removed.
\end{itemize}

\subsubsection{Human evaluation}
To check the validity of the test set, the first author manually examined 160 generated nonce sentences in the same fashion that the model would evaluate them.
That is, each example comprised either a singular or plural noun inserted into a template, and the first author had to predict the correct number inflection of a given verb. 
In addition, the first author had to verify that the sentence was grammatical and contained no number inflection cues other than the inflection of the subject.
The sentences were presented in random order, with the first author blinded from the ground-truth label.

The first author found that 6 sentences (3 templates, since each template had two inflections) contained additional number inflection cues that were missed in the first round of annotation, and so these sentences were removed.
In terms of accuracy, the first author correctly predicted the verb inflection 153 of the 154 instances (99.4\% accuracy) and attributed the single error to carelessness. 
We take these manual evaluation results as evidence that our test set is grammatical and tests a syntactic rule that can be consistently applied by humans.

\newpage 
\section{Additional Figures and Tables}
We show several auxiliary experiments that elucidate BERT's performance with respect to various characteristics of evaluation stimuli. 

\subsection{Relative Frequency of Forms}
\begin{figure}[ht]
    \centering \includegraphics[width=0.75\linewidth]{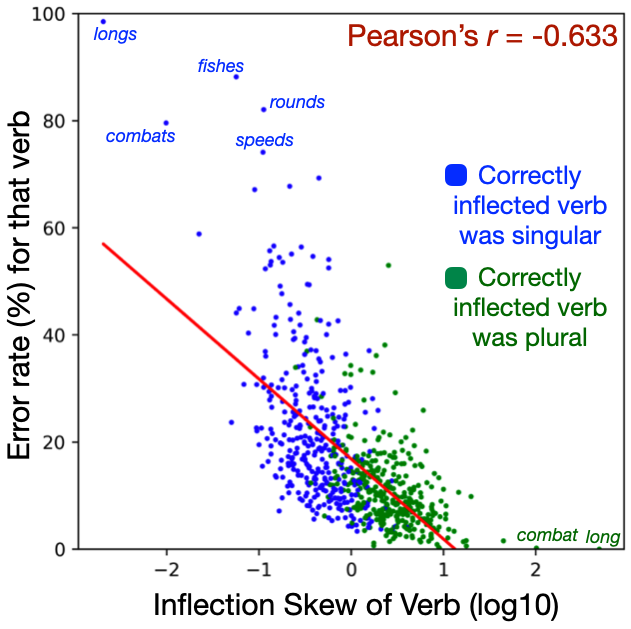}
    \vspace{-2mm}
    \caption{ For all 366 verbs ($\times2$ inflections per verb), we plot the error rate against \textit{inflection skew}, which is how much more frequent the correct inflection of the verb occurred than the incorrect inflection in the pre-training data. 
    Several of the most skewed words are shown---for instance, \textit{long} appears 490 times more often in the pre-training data than \textit{longs}, so its inflection skew is $\log_{10}(490) = 2.69$. Conversely, \textit{longs} has an inflection skew of $-2.69$. 
    } \label{fig:verb_inflection_asymmetry_points}
\end{figure}

Following the results from \cref{subsec:ratio_frequency}, we show the error rate for singular and plural forms of all verbs in our nonce stimuli in \cref{fig:verb_inflection_asymmetry_points}. 
Additionally, \cref{tab:verb_inflection_asymmetry_examples} shows the five verbs with the highest and lowest error rates, as well as their frequency ratios.

\subsection{Comparison with prior work}
Because our work proposes a new set of nonce stimuli (which is larger than prior work, e.g., 383 sentences from \citet{gulordava-etal-2018-colorless} or 7 verbs from \citet{chaves-2021-look}), we run several analyses from prior work on our dataset. 
The results are largely consistent with conclusions from prior work. 

\paragraph{Attractors.}
As shown in \cref{tab:templates_attractors}, BERT did not perform worse on templates with more attractors (clauses between the subject and verb), corroborating \citet{goldberg2019assessing}. 

\begin{figure}[ht]
    \centering \includegraphics[width=0.75\linewidth]{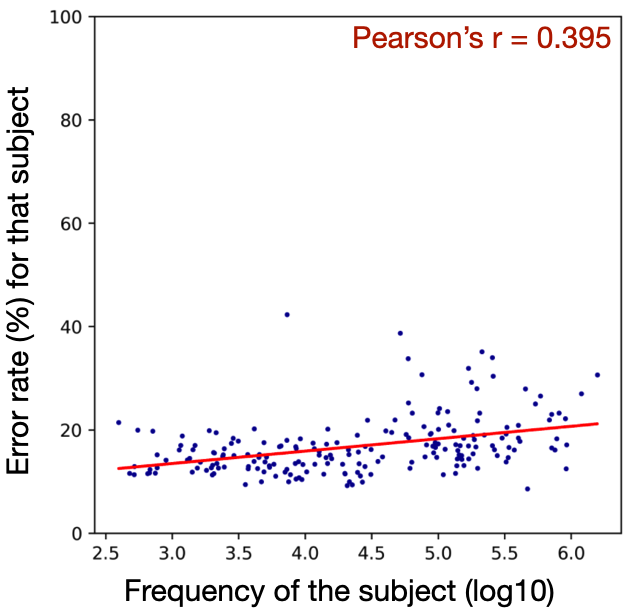}
    \caption{
    For all 200 subjects (both singular and plural forms are included into a single subject), we plot the error rate against the frequency of that subject in the pre-training data.
    } \label{fig:subj_freq_main}
\end{figure}

\paragraph{Noun frequency.}
We find similar evidence, like \citet{yu-etal-2020-word}, that BERT did not perform better on nouns that were more frequent in the training set.
\cref{fig:subj_freq_main} shows these results---for each subject (we consider both singular and plural forms as a single subject), we plot that subject's error rate against its frequency in the training data.

\begin{figure}[ht]
    \centering
    \includegraphics[width=0.75\linewidth]{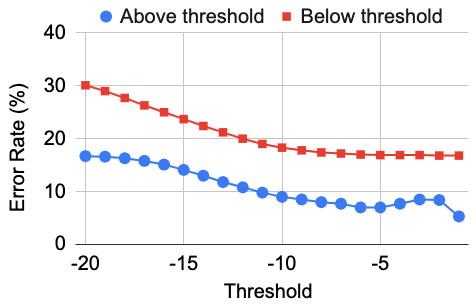}
    \caption{
    Error rates of examples for which the model's predictions were above and below certain thresholds. 
    }
    \label{fig:confidence_graph}
\end{figure}

\paragraph{High- and low-confidence predictions.}
As an auxiliary analysis to compare with \citet{newman-2021-refining}, we plot the error rate of BERT with respect to different thresholds for how confident the model was about its prediction.
\cref{fig:confidence_graph} shows error rates for all predictions with confidence above some threshold, and error rates for predictions with confidence below some threshold.
This result concurs with \citet{newman-2021-refining}'s finding that model performance drops when testing verbs that the model finds unlikely.

\subsection{Absolute versus relative frequency}\label{appendix:abs_vs_relative}
\begin{figure}[ht]
    \centering \includegraphics[width=0.75\linewidth]{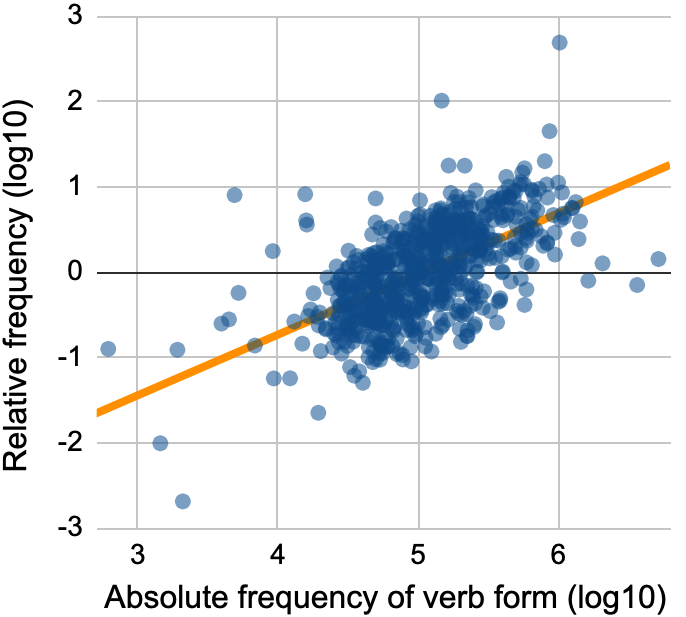}
    \caption{
    For all 336 verbs (672 verb forms), we plot the absolute frequency of a verb form versus the relative frequency of that verb form compared with its competing form. 
    Pearson's $R^2 = 0.356$.  
    } \label{fig:abs_vs_relative}
\end{figure}

As additional background, \cref{fig:abs_vs_relative} shows the correlation between absolute frequency of a verb form and its frequency relative to its competing form.

\begingroup
\begin{table}[h]
    \centering
    \small
    \begin{tabular}{l | c c}
    
        \toprule
         & Inflection Skew \\
         & (Log10) & Error Rate \\
        \midrule
        \multicolumn{1}{l}{Best-performing verbs} \\
        long & 2.69 & 0.0\% \\
        speed & 0.95 & 0.2\% \\
        combat & 0.95 & 0.2\% \\
        round & 2.01 & 0.4\% \\
        fish & 1.25 & 0.6\% \\
        \midrule
        \multicolumn{3}{l}{Worst-performing verbs} \\
        longs & -2.69 & 98.5\% \\
        fishes & -1.25 & 88.0\% \\
        rounds & -0.95 & 82.1\% \\
        combats & -2.01 & 79.6\% \\
        speeds & -0.95 & 74.1\% \\
        \bottomrule
    \end{tabular}
    \caption{
    The verbs for which BERT had the highest and lowest error rates. 
    Inflection skew indicates how much more often a verb appeared in BERT's pre-training data compared with its other number inflection, in log10. 
    For instance, an inflection skew of 1 indicates that a verb appeared $10^1=10$ times more often in the training set than its other number inflection.
    }
    \label{tab:verb_inflection_asymmetry_examples}
\end{table}
\endgroup
\begingroup
\begin{table}[ht]
    \centering
    \small
    \begin{tabular}{p{4cm} | c c}
        \toprule
        Stratification of Stimuli & Examples & Error Rate \\
        \midrule
        All examples & 10.2M & 17.9\% \\
        Templates with one attractor & 3.9M & 15.2\% \\
        Templates with two attractors & 1.6M & 22.3\% \\
        Templates with three attractors & 1.3M & 16.8\% \\
        Templates with four attractors & 672k & 13.0\% \\
        \bottomrule
    \end{tabular}
    \caption{
    Performance for templates with different numbers of attractors (distracting clauses between the subject and verb).
    }
    \label{tab:templates_attractors}
\end{table}
\endgroup

\clearpage 
\subsection{Training manipulations: Seen vs.\ Unseen}\label{appendix:seen_vs_unseen}
\cref{fig:mani_raw_freq} in the main body showed the performance of models that have seen the VOI $n$ times in training, where $n$ varies from 1 to 10,000. 
\cref{tab:stratify_seen_unseen} stratifies this performance on the nonce evaluation stimuli by seen and unseen subject--verb pairs in the evaluation stimuli. 
Note, though, that this stratification differs for each VOI frequency.
That is, there will be more unseen subject--verb pairs when the VOIs are less frequent in the training set. 
\begingroup
\setlength{\tabcolsep}{3pt}
\begin{table}[ht]
    \centering
    \small
    \begin{tabular}{l | c c | c c}
        \toprule
        Frequency & \multicolumn{2}{c|}{Seen SV} & \multicolumn{2}{c}{Unseen SV} \\
        of VOI & \# examples & Error & \# examples & Error \\
        \midrule
        1 & 672 & 63.4\% & 1.34M & 50.1\% \\
        3 & 2.6k & 54.8\% & 1.34M & 48.9\% \\
        10 & 7.4k & 43.7\% & 1.34M & 45.1\% \\
        30 & 21.6k & 29.6\% & 1.32M & 30.8\% \\
        100 & 59.9k & 20.9\% & 1.28M & 23.7\% \\
        300 & 135k & 22.1\% & 1.21M & 22.8\% \\
        1,000 & 289k & 20.4\% & 1.06M & 20.0\% \\
        3,000 & 456.3k & 16.3\% & 887.7k & 16.9\% \\
        10,000 & 662.k & 15.4\% & 682.1k & 14.5\% \\
        \bottomrule
    \end{tabular}
    \caption{
    Stratifying seen (frequency > 0) and unseen (frequency = 0) subject--verb pairs in the nonce sitmuli for our training manipulation from \cref{fig:mani_raw_freq}.
    }
    \label{tab:stratify_seen_unseen}
\end{table}
\endgroup

\clearpage 
\section{Raw SVA Nonce Stimuli}\label{appendix:raw_sva_dataset}
\subsection{Verbs}
The 336 verbs used in our nonce stimuli are listed below (only plural/base inflections are shown):
\noindent {\small
add, 
advance, 
age, 
aim, 
air, 
allow, 
analyse, 
angle, 
approach, 
archive, 
arrive, 
ask, 
assist, 
attack, 
attempt, 
award, 
bar, 
base, 
battle, 
bear, 
become, 
begin, 
believe, 
benefit, 
block, 
board, 
bond, 
book, 
border, 
branch, 
brand, 
break, 
bridge, 
bring, 
call, 
campaign, 
carry, 
cause, 
center, 
centre, 
challenge, 
champion, 
change, 
channel, 
charge, 
chart, 
circle, 
claim, 
class, 
coach, 
code, 
color, 
colour, 
combat, 
comment, 
compound, 
comprise, 
concern, 
connect, 
consider, 
contact, 
contain, 
continue, 
contract, 
control, 
copy, 
count, 
course, 
cover, 
create, 
credit, 
critique, 
crop, 
cross, 
cycle, 
date, 
deal, 
debate, 
decide, 
define, 
demand, 
describe, 
design, 
detail, 
develop, 
discover, 
display, 
dispute, 
distance, 
document, 
double, 
draw, 
drive, 
drug, 
effect, 
end, 
enter, 
equip, 
estimate, 
exhibit, 
experience, 
explain, 
extend, 
eye, 
face, 
factor, 
fan, 
farm, 
feature, 
feel, 
field, 
fight, 
file, 
fill, 
finance, 
find, 
fire, 
fish, 
fly, 
follow, 
force, 
form, 
frame, 
fund, 
gain, 
get, 
give, 
grade, 
graduate, 
grant, 
group, 
grow, 
guard, 
guide, 
hand, 
have, 
head, 
help, 
hold, 
honor, 
honour, 
host, 
house, 
include, 
increase, 
indicate, 
influence, 
interview, 
involve, 
issue, 
join, 
judge, 
keep, 
kill, 
know, 
label, 
land, 
lap, 
lead, 
learn, 
leave, 
level, 
light, 
limit, 
line, 
link, 
list, 
live, 
long, 
look, 
love, 
maintain, 
make, 
manage, 
map, 
mark, 
market, 
master, 
match, 
matter, 
mean, 
measure, 
meet, 
mention, 
minister, 
model, 
move, 
murder, 
name, 
need, 
note, 
number, 
object, 
offer, 
open, 
operate, 
order, 
own, 
pair, 
park, 
partner, 
pass, 
pattern, 
pay, 
peak, 
perform, 
picture, 
pilot, 
place, 
plan, 
plant, 
play, 
position, 
post, 
pound, 
power, 
practice, 
present, 
print, 
process, 
produce, 
program, 
project, 
protest, 
prove, 
provide, 
question, 
race, 
raid, 
range, 
rank, 
rate, 
reach, 
reason, 
receive, 
record, 
refer, 
reference, 
reflect, 
refuse, 
release, 
remain, 
repair, 
report, 
represent, 
reprise, 
require, 
reserve, 
return, 
reveal, 
review, 
ring, 
rise, 
risk, 
rival, 
round, 
route, 
rule, 
run, 
sample, 
say, 
scale, 
score, 
seat, 
see, 
seed, 
seek, 
send, 
serve, 
service, 
share, 
ship, 
show, 
sign, 
signal, 
single, 
sketch, 
slope, 
sound, 
source, 
speak, 
speed, 
sport, 
spot, 
stage, 
stand, 
star, 
start, 
state, 
stop, 
store, 
stream, 
strike, 
strip, 
structure, 
study, 
style, 
suggest, 
supply, 
support, 
surface, 
tackle, 
take, 
talk, 
target, 
task, 
tax, 
tell, 
term, 
test, 
tour, 
trace, 
track, 
trail, 
train, 
transport, 
travel, 
trend, 
trouble, 
try, 
turn, 
use, 
value, 
vary, 
vent, 
view, 
visit, 
voice, 
volunteer, 
walk, 
want, 
wave, 
win, 
witness, 
work, 
write, 
}

\subsection{Nouns}
The 200 nouns used in our nonce stimuli are listed below (only singular inflections are shown): 
\noindent {\small
abuser, 
actuary, 
affiliate, 
album, 
application, 
artefact, 
articulation, 
artiste, 
aspect, 
astronomer, 
attempt, 
attribution, 
autopilot, 
ball, 
barnacle, 
basalt, 
batch, 
battalion, 
beaker, 
bettor, 
bidder, 
biosensor, 
blazer, 
bluebird, 
brake, 
brush, 
bulletin, 
business, 
campaign, 
capital, 
captor, 
caretaker, 
catholic, 
caveman, 
charge, 
chestnut, 
clarinetist, 
climate, 
columnist, 
command, 
commando, 
commuter, 
comparison, 
compiler, 
constant, 
consul, 
craftsman, 
credential, 
cup, 
debate, 
debater, 
demigod, 
device, 
dhole, 
disorder, 
distribution, 
diviner, 
draft, 
drum, 
dynasty, 
echidna, 
electron, 
emoticon, 
enclave, 
etymology, 
exhibition, 
explosive, 
faith, 
fanatic, 
fantasy, 
fat, 
ferret, 
fiction, 
foal, 
forager, 
form, 
forwarder, 
fossil, 
foundation, 
franchise, 
friendship, 
girl, 
glass, 
good, 
grantee, 
grapevine, 
hair, 
harmonic, 
headlamp, 
hedgehog, 
hotel, 
hypothesis, 
imp, 
impact, 
instruction, 
intensifier, 
interest, 
intrusion, 
island, 
isometry, 
kabbalist, 
kind, 
launch, 
layer, 
legionnaire, 
lioness, 
loading, 
locksmith, 
logarithm, 
logger, 
mammoth, 
martin, 
matchup, 
microphone, 
misfit, 
motorcyclist, 
nasal, 
necessity, 
officer, 
ogre, 
opposition, 
palace, 
panchayat, 
parrot, 
pioneer, 
platform, 
plum, 
poet, 
possibility, 
postposition, 
potentiometer, 
president, 
press, 
pro, 
proponent, 
provider, 
race, 
radiologist, 
rank, 
rat, 
reaper, 
region, 
relief, 
remark, 
repeater, 
repellent, 
rescuer, 
researcher, 
retriever, 
ribbon, 
ride, 
ring, 
rogue, 
role, 
sage, 
salaryman, 
seagull, 
section, 
selection, 
sense, 
sex, 
shearer, 
sheepdog, 
shoreline, 
siding, 
sign, 
simulation, 
situation, 
skateboarder, 
snowflake, 
sorcerer, 
specimen, 
speech, 
spill, 
spiritualist, 
spore, 
spring, 
starling, 
starship, 
stingray, 
stock, 
street, 
suffix, 
switch, 
tarsier, 
terrier, 
town, 
treaty, 
truth, 
tutor, 
tweeter, 
undertaker, 
uniform, 
vendor, 
ventilator, 
view, 
walker, 
warlock, 
watcher, 
youngster}

\subsection{Sentential Contexts}
The 56 sentential contexts used in our stimuli are listed below:
\begin{enumerate}[leftmargin=*]
\small
\itemsep0em 
    \item the edwardian semi-detached [SUBJECT] of brantwood road , facing the park [VERB] an art deco style whilst those in ashburnham road include ornate balconies . \item
for protestant denominations , the [SUBJECT] of marriage [VERB] intimate companionship , rearing children and mutual support for both husband and wife to fulfill their life callings . \item
the [SUBJECT] , due to being the same colour green as the shield , [VERB] a green sign with a white inlay border , and a green outer border . \item
wright 's other acting [SUBJECT] on television [VERB] itv 's crossroads and bbc one 's doctors . \item
the [SUBJECT] , where most of the population lives and the majority of activity takes place , [VERB] an expanse of low-lying , flat , and comparatively dry grassland . \item
the [SUBJECT] of the load line with the transistor characteristic curve [VERB] the different values of ic and vce at different base currents . \item
the other [SUBJECT] on the pillar [VERB] only two bolts for sport climbing , making a ground-fall more likely should a mistake be made . \item
the [SUBJECT] honoring saint joseph , the saint patron of the city , [VERB] a part of the city 's culture . \item
furthermore , other scholars have noted how the cryptic dharani [SUBJECT] within the lotus sutra [VERB] a form of the magadhi dialect that is more similar to pali than sanskrit . \item
he may be a wonderful vandal fighter , but i think the [SUBJECT] about this matter at the talk page [VERB] a clear misunderstanding of practice , here . \item
bce ) , although no [SUBJECT] of that period [VERB] today . \item
its major american [SUBJECT] in the plastic model kit market [VERB] amt-ertl , lindberg , and testors . \item
his feature [SUBJECT] as a screenwriter [VERB] american hot wax , rafferty and the gold dust twins and where the buffalo roam . \item
hence , some state [SUBJECT] of assault weapon explicitly [VERB] assault rifles . \item
his other research [SUBJECT] in modern cornish history [VERB] cornish emigration ; ethnicity and territorial politics and centre-periphery relations . \item
her [SUBJECT] of interest [VERB] the history of childhood and family , networks , social interactions and reciprocity , poverty , welfare , gift-exchange and the history of the emotions . \item
the [SUBJECT] for approval of an employment visa [VERB] suitable educational qualifications or work experience , a secured employment contract in moldova , provide proof of adequate means of subsistence in moldova , police confirmation that you have no criminal record , and a satisfactory medical examination . \item
the [SUBJECT] of this measure [VERB] not a single reason to advance why this bill should not pass . \item
the control [SUBJECT] at the left hand end of the instrument [VERB] the d6 clavinet mixture controls and a sliding control for the volume . \item
second of all , his [SUBJECT] today [VERB] very high prices , placing him well above the notability minimum for artists . \item
communities bans should be the absolute last resort when an editor 's [SUBJECT] to the project [VERB] a net detrimental effect and lesser sanctions have failed to improve this problem . \item
the [SUBJECT] of the first session of the washington county court during that year [VERB] a call for a road from canon 's mill to pittsburgh . \item
the anaerobic [SUBJECT] in osteomyelitis associated with peripheral vascular disease generally [VERB] the bone from adjacent soft-tissue ulcers . \item
i have taken note of michaelqschmidt 's keep vote , but note that the [SUBJECT] in the second google news link [VERB] nothing non-trivial and the first shows only brief local coverage . \item
the small [SUBJECT] of non-white students in the schools accurately [VERB] the racial and ethnic demographics of the community . \item
however , the [SUBJECT] of cost versus benefit [VERB] an area of ongoing research and discussion . \item
the common [SUBJECT] for dizziness [VERB] vertigo , pre-syncope and disequilibrium . \item
the [SUBJECT] of the increasing lack of physical education [VERB] budgetary pressure which limits the resources provided for this ; the increasing attractiveness of rival pastimes such as video games ; and the increased emphasis upon academic results . \item
the [SUBJECT] of the former route from drysdale to south geelong , along with a walking track adjacent to the queenscliff-drysdale line , now [VERB] the bellarine rail trail , accessible to cyclists and walkers . \item
one 's [SUBJECT] at this level of existence [VERB] a consistency and coherence that they lacked in the previous sphere of existence . \item
the [SUBJECT] in the gulf [VERB] santa catalina island . \item
the oaths themselves talk about the family bond , and we can conjecture that the [SUBJECT] of secrecy [VERB] the family loyalty as well as a sense of self-preservation . \item
the [SUBJECT] on death row [VERB] foreign nationals , many of whom were convicted of drug-related offences . \item
the [SUBJECT] on current routes [VERB] nothing to the info . \item
the [SUBJECT] of the buildings [VERB] commercial space , including two restaurants , a dental office ( pinnacle dental ) , a medical clinic , and spa , while the surrounding area will consist of public parks , shops and recreation spaces . \item
the genetic [SUBJECT] among the viruses isolated from different places ( 7-8 ) [VERB] the difficulty of developing vaccines against it . \item
the [SUBJECT] of shipwrecks in 1980 [VERB] all ships sunk , foundered , grounded , or otherwise lost during 1980 . \item
the [SUBJECT] of these techniques to humans [VERB] moral and ethical concerns in the opinion of some , while the advantages of sensible use of selected technologies is favored by others . \item
under chairwoman agnes gund , the moma ps1 's [SUBJECT] of directors [VERB] the artists laurie anderson and paul chan , art historian diana widmaier-picasso , fashion designer adam kimmel , and art collectors richard chang , peter norton , and julia stoschek . \item
the first [SUBJECT] of `` cities of the plain '' [VERB] a detailed account of a sexual encounter between m . \item
the diocese 's [SUBJECT] of arms [VERB] a red field in honor of the sacred heart of jesus . \item
the [SUBJECT] from local schools , like west lafayette junior-senior high school , also [VERB] high academic standards . \item
the [SUBJECT] of chaperones [VERB] a long history . \item
his [SUBJECT] in rabies [VERB] multiple studies investigating efficacy and side effects of tissue culture derived rabies vaccines , as well as leading clinical trials as primary investigator in collaboration with the who . \item
the [SUBJECT] of glaciers on people [VERB] the fields of human geography and anthropology . \item
the object is to eliminate as many stars as possible before the [SUBJECT] of blocks [VERB] the top of the screen ; a hand raises up the set of blocks , introducing a new row . \item
the [SUBJECT] of all preordered sets with monotonic functions as morphisms [VERB] a category , ord . \item
the [SUBJECT] of goods and services chosen [VERB] changes in society 's buying habits . \item
the [SUBJECT] of binding partners to induce conformational changes in proteins [VERB] the construction of enormously complex signaling networks . \item
the [SUBJECT] of university of toledo people [VERB] notable alumni , former students , and faculty of the university of toledo . \item
romayne 's film [SUBJECT] scoring independent features and documentaries [VERB] the screamfest crystal skull winner h . \item
the [SUBJECT] of basic needs providers emigrating from impoverished countries [VERB] a damaging effect . \item
the [SUBJECT] of extensive deposits of fishbones associated with the earliest levels also [VERB] a continuity of the abzu cult associated later with enki and ea . \item
in addition , the [SUBJECT] of secondary measures [VERB] applications in quantum mechanics . \item
the [SUBJECT] of large quantities [VERB] specific precautions to prevent the release of the vapour into the environment . \item
the [SUBJECT] with the highest votes [VERB] the deputy mayor and may proxy for the mayor .
\end{enumerate}

\subsection{Verbs of Interest}\label{appendix:voi}
The 60 verbs of interest (VOI) used in our pre-training manipulation experiments are listed below:
{\small
emphasize,
threaten,
announce,
utilize,
propose,
translate,
confront,
portray,
prefer,
declare,
denote,
admit,
conclude,
inform,
imply,
relate,
derive,
suffer,
constitute,
employ,
possess,
attract,
assume,
resemble,
depict,
demonstrate,
incorporate,
celebrate,
generate,
realize,
collect,
enjoy,
deliver,
introduce,
explore,
prepare,
depend,
recognize,
encourage,
contribute,
hear,
publish,
retain,
discuss,
enable,
prove,
spend,
comprise,
define,
marry,
affect,
teach,
argue,
survive,
choose,
identify,
lose,
vary,
raise,
reveal
}

\end{document}